\documentclass[11pt,a4paper]{article}
\usepackage[hyperref]{acl2021}
\usepackage{times}
\usepackage{booktabs}
\usepackage{todonotes}
\usepackage{latexsym}
\usepackage{float}
\usepackage{bbding}
\usepackage{pifont}
\usepackage{wasysym}
\usepackage{amssymb}
\usepackage{multirow}
\usepackage{makecell}
\usepackage{subcaption}

\usepackage{microtype}

\aclfinalcopy

\title{Bridging the Gap: Exploring the Capabilities of Bridge-Architectures for Complex Visual Reasoning Tasks}

\author{
  Kousik Rajesh\thanks{\hspace{4pt}Everyone Contributed Equally -- Alphabetical order} \hspace{2em} Mrigank Raman$^*$ \hspace{2em} Mohammed Asad Karim$^*$ \hspace{2em} Pranit Chawla$^*$ \\
  \texttt{\{kousikr, mrigankr, mkarim2, pranitc\}@andrew.cmu.edu}
  }

\date{}

\begin{document}
\maketitle
\begin{abstract}
In recent times there has been a surge of multi-modal architectures based on Large Language Models, which leverage the zero shot generation capabilities of LLMs and project image embeddings into the text space and then use the auto-regressive capacity to solve tasks such as VQA, captioning, and image retrieval. We name these architectures as "bridge-architectures" as they project from the image space to the text space. These models 
deviate from the traditional recipe of training transformer based multi-modal models, which involve using large-scale pre-training and complex multi-modal interactions through co or cross attention. However, the capabilities of bridge architectures have not been tested on complex visual reasoning tasks which require fine grained analysis about the image. In this project, we investigate the performance of these bridge-architectures on the NLVR2 dataset, and compare it to state-of-the-art transformer based architectures. We first extend the traditional bridge architectures for the NLVR2 dataset, by adding object level features to faciliate fine-grained object reasoning. Our analysis shows that adding object level features to bridge architectures does not help, and that pre-training on multi-modal data is key for good performance on complex reasoning tasks such as NLVR2. We also demonstrate some initial results on a recently bridge-architecture, LLaVA, in the zero shot setting and analyze its performance.
\end{abstract}

\section{Introduction and Problem Definition }
With the advent of transformer based architectures \cite{bert, roberta} for natural language processing tasks, these architectures were naturally extended for multi-modal use cases, by treating both image regions and text tokens as joint tokens and perform self and cross attention across them \cite{lxmert, uniter, visbert, vilbert, xvlm}. However, these models followed a pre-train and fine-tune strategy, where they were first pre-trained using image-text matching and masked language modelling on very large scale image captioning datasets such as CC12M \cite{cc12m}, MSCOCO \cite{mscoco}. For instance X-VLM and UNITER were pre-trained on close to 15 million pairs of image and text. This pre-training step is both computationally and data intensive. Moreover, even after pre-training, these architectures were required to be fine-tuned on specific tasks such as VQA \cite{VQA}, \cite{applications_image_retrieval}, \cite{caption1}. This results in both the redundancy of weights, since each model needs to be trained on every dataset explicitly, and also implies that such architectures cannot be used in a zero shot setting. Moreover, UNITER, and X-VLM need explicit cross-modal attention across the image and text tokens, which is often computationally expensive. 
In recent times, LLMs \cite{opt, gpt} have shown extremely impressive results on zero shot language reasoning tasks, and these architectures were again extended to work in the multi-modal setting \cite{fromage, magma, frozen, limber} by using a pre-trained vision encoder such as CLIP \cite{clip} to obtain image embeddings and then project them into the text space, followed by using an LLM such as \cite{opt, raffel2020exploring} to decode the embeddings to generate text and then use this generated text to perform downstream tasks, either in the zero-shot or few-shot settings. We name these architectures as the "bridge architectures". These architectures have a lot of adavantages over the traditional transformer based ones. First, they do not require to be pre-trained on very large scale multi-modal data, as they use pre-trained image and text encoders. Second, since the output of such models is in natural language, their reasoning is much easier to understand compared to the earlier ones. Moreover, brige architectures do not have explicit cross-modal attention blocks, which makes then computationally cheaper. Finally, since these architectures leverage the zero-shot reasoning capabilities of LLMs, they often do not need to be fine-tuned on specific datasets. 
However, the performance bridge architectures has not been investigated on complex visual reasoning tasks such as NLVR2, which we believe is an extremely vital benchmark to judge the multi-modal reasoning capabilities of models. Thus, in this project, we analyze the performance of such architectures on the NLVR2 dataset. Since NLVR2 requires extremely complex reasoning across multiple images and objects, we find the current bridge architectures lacking since they all use a global feature vector to represent the image, which would often be insufficient for NLVR2. Thus, we add pre-trained object features from a model such as \cite{fast_rcnn, sam} to the global feature vector and then follow the same architecture as traditional bridge architectures. 
Through our experiments, qualitative and quantitative analyses, we try to answer the following three \textbf{research questions}:
\textit{\begin{itemize}
    \item \textbf{RQ1:} Does incorporating object level local features help bridge architectures in this task?
    \item \textbf{RQ2:} Is visuo-linguistic pretraining necessary for good performance on multimodal tasks?
    \item \textbf{RQ3:} Is multimodal instruction finetuning neccesary for bridge architectures?
\end{itemize}}

Our contributions can be summarized as follows:
\begin{itemize}
    \item We show results on incorporating object level local features from SAM~\cite{sam} which is a segmentation model into a vanilla bridge architecture without multimodal pretraining. Our results indicate that local features dont help much at all unless the bridge architecture is pretrained. Due to resource constraints, we were unable to pretrain a bridge architecture with SAM and we address it as future work. 

    \item Our analysis of various multimodal models reveal that it is imperative that the models are pretrained on visuo-linguistic data. Our evaluations also hint to the fact that pretraining the image encoder is very important to perform well on NLVR2

    \item Our experiments with LLaVA demonstrate that instruction finetuning is very important for leveraging the Chain of Thought (CoT) reasoning of the LLM which helps the zero-shot performance quite a lot. We posit that finetuning the image encoder via task instruction would be very beneficial but due to resource constraints, we leave it a future work.
\end{itemize}

\section{Related Work and Background }
\paragraph{Related Datasets} 
One of the ealiest datasets for VQA \cite{suhr2017corpus} by Agrawal et al. suffered from severe biases.  These biases allowed unimodal baselines to achieve performance close to the state of the art. A unimodal baseline \cite{GoyalKSBP16} that used solely the language data achieved an accuracy of 48\% when the state of the performance was at 61\%.\\
An extension to the dataset (VQA2.0) that fixed some of the biases \cite{GoyalKSBP16} was proposed. The authors collect complementary images such that each question can have two possible answers from the training data.
\\
Although such methods attempted to mitigate biases present in VQA datasets, it is difficult to apply such bias-reduction approaches
to more complex questions without a high-quality semantic representation of both questions and answers.
The CLEVR dataset \cite{johnson2017clevr} provides a dataset that requires complex reasoning to solve, this level of control over the dataset is achieved by synthetically generating images and questions.
The NLVR dataset \cite{suhr-etal-2017-corpus} also contains synthetically generated images but crowdsourced questions. This questions and sentences involve reasoning about multiple objects and attributes as compared to CLEVR. The questions are also more ambiguous and require subtle reasoning to answer. \\
The NLVR2 dataset \cite{nlvr2} was introduced for joint reasoning about natural language and images, with a focus on semantic diversity, compositionality,
and visual reasoning challenges. Using real photographs instead of synthetically generated ones requires models to engage in more abstract reasoning and common sense knowledge to solve the task. The dataset also makes an attempt to remove biases present in the original photograph based VQA datasets by carefully choosing sysnsets and answers. \\
Further attempts to improve diversity of language in VQA datasets was done by \cite{marvl} in the Multicultural Reasoning
over Vision and Language (MaRVL) dataset. The authors use a similar dataset generation methodology to NLVR2, over five typologically diverse languages - Indonesian, Mandarin Chinese, Swahili, Tamil, and Turkish. They incorporate language specific concepts by eliciting topics from native speaker annotators and making them annotate any statements about pairs of images. The authors acknowledge that some biases may still exist in the dataset due to the relatively low number of annotators available for the low resource languages.

\subsection{Prior Work}

\paragraph{UNITER}\cite{10.1007/978-3-030-58577-8_7} Joint-text and visual embedding is one of the key aspects of multimodal learning. The proposed model aims to learn the joint text and visual representations via several pre-training tasks (Masked Language Modeling (MLM), Masked Region Modeling (MRM, with three variants), ImageText Matching (ITM), and Word-Region Alignment (WRA) over several multimodal datasets. Tasks such as Word-Region Alignment are very relevant to NLVR2 given the task complexity.

\paragraph{SimVLM}\cite{Wang2021SimVLMSV} Methods that leverage large-scale vision and language pretraining tasks to learn representations for downstream tasks generally need annotations while pretraining. SimVLM alleviates this issue by using large-scale weak supervision and is trained on the Prefix Language Modelling Task. Even with reduced dataset complexity, SIMVLR is able to outperform state of art methods on several vision and language tasks. Large-scale supervision improves generalization and transfer abilities. Such characteristics are critical for reasoning tasks like NLVR2.

\paragraph{ALBEF}\cite{NEURIPS2021_50525975} 
Existing methods in large-scale vision and text representation learning use a transformer-based model to jointly encode visual and language representations. Such encoding is challenging because the visual and text embedding space are not aligned. Albef uses a contrastive loss followed by cross-modal attention to tackle this problem. Such a method does not require annotations like the bounding boxes.

\paragraph{SOHO}\cite{Huang_2021_CVPR} State of art cross-alignment methods first extract important regions from the image and then align them with their corresponding words. Due to this, vision-language model fail to capture the complete semantics from natural language. SOHO takes the complete image as input and produces semantically relevant vision language representations. SOHO leverages a visual dictionary to generate image features that facilitate cross-modal understanding. SOHO outperforms other methods on visual-language tasks by a significant margin. To the best of our knowledge, we would be the first to analyze Language Models on this dataset. SOHO's results on dev and test-p split for NLVR2 demonstrates the efficacy of the method on compositional reasoning tasks.

\subsection{Relevant techniques} Most of the earlier work in vision-language understanding focused on using multi-modal transformers such as ViLT \citep{vilt}, UNITER \citep{uniter}, VisualBERT \citep{visbert} to align the image and text representations to a common space using self and cross attention, and then fine tuning these joint representations for a specific downstream task. However, such multi-modal representations are hard to interpret and it is difficult for users to interact with such models. Recently, with the advent of Large Language Models \cite{llama, gpt, opt}, these models have achieved extremely impressive performance on zero-shot and few-shot language generation and reasoning tasks, and at the same time, have provided users and researchers with a way to understand their predictions and decoding processes, have allowed researchers to prompt and provide chain of thoughts to these models to improve their performance \cite{chain_thought}, and have given a way to add facts/concepts to the model without changing its weights \cite{edit}. Due to these advantages, recent works such as Frozen \citealp{frozen} and MAGMA \citep{magma} demonstrated the efficacy of large language models on few shot multi-modal reasoning and captioning tasks, by allowing LLMs to take as input sequences of images and text. They do this by  feeding embeddings of images (obtained by an encoder such as \cite{clip, resnet}) through a frozen LLM, and back-propogating through the weights of the image encoders, to project the tokens into the input space of LLMs. While \cite{frozen, magma} achieve impressive results, one bottle-neck is still the end-to-end pre-training of the image encoder, which is expensive. To alleviate this issue, \cite{limber, fromage} showed that embeddings from image encoders can be linearly projected to the space of LLMs by a training a single projection matrix, while keeping the two encoders frozen. We find this direction to be promising and in this project, we want to focus on building onto such approaches of learning linear mappings from images to text for the NLVR2 dataset. We describe the exact architecture we use for NLVR2 in the next section. 
One challenge for this task is that most examples in our dataset are reliant on local image features, such as the count, position or locations of objects, which aren't encoded very well by existing image encoders. We plan to leverage object level and positional information by using the embeddings of \cite{fasterrcnn}, adding positional information as done in \cite{lxmert}, and then projecting them to the text space. Moreover, since the initial objects obtained from \cite{fasterrcnn} are usually trained only on image data, it might be a little hard to project them directly into the text space by a linear matrix. Another direction we are planning to look into to improve the multi-modal interaction between the local image features and input sentence is to use end-to-end transformer based language grounders \citealp{mdetr, transvg}, which return referred objects and have been trained with language supervision. Using global features from \cite{clip} and local features from \cite{mdetr} might help us better align the images to text space, by leveraging the language supervision used in the training of both the models. One week ago we came across LLaVA~\cite{llama}, a bridge architecture that has been pretrained on cc3m~\cite{cc3m} data and finetuned on multimodal task instructions. We use LLaVA in our detailed analysis. The authors of \citet{cot} demonstrate that an LLM's zero shot capabilities can be enhanced by using Chain of Thought reasoning. In this analysis, we try to elicit complex zero-shot reasoning of instruction finetuned and pre-trained LLMs by using CoT reasoning on LLaVA. In this study, we also try bootstrapping LLaVA using its own CoT reasoning in the same vein as \citet{selfimprove}. Along with LLaVA, we also include other bridge architectures such as Fromage~\cite{fromage} and Open-Flamingo~\cite{open-flamingo}.

\section{Task Setup and Data}
We worked on the NLVR2 dataset~\citep{suhr2018corpus}. Given a pair of images and a text, the task is to predict whether the given caption about the pair of images is true/false. This dataset tests the ability to align text and pair of images and jointly reason over them which is easy for humans but very difficult for machines. To complicate matters for machines, this dataset also has cardinality-based captions where the task is to compare the number of objects between two images and reason about it via text.

\section{Baselines}
\begin{table*}[ht]
\centering
\begin{tabular}{@{}lrrrr@{}}
\toprule
                            & \multicolumn{3}{c}{Dev} \\
Methods                     & Accuracy $\uparrow$ & Recall $\uparrow$ & Precision $\uparrow$ & F1 Score $\uparrow$  \\
\midrule
RoBERTa (Text Only) \cite{roberta} & 50.85 & 100.00 & 50.85&67.43\\
ResNet-50 (Image Only) \cite{resnet} & 51.92 & 49.14& 52.94&50.97 \\
MAE (Image Only) \cite{mae} & 51.67 & 54.29& 52.40& 53.33\\
\midrule
MAE + RoBERTa  & 52.00 & 72.26& 51.90& 60.41\\
ViLT-LP \cite{vilt} & 62.10 & 72.23& 60.71& 65.97\\
X-VLM-LP \cite{xvlm} & 69.90 & 70.38 & 69.76 & 69.62 \\
\midrule
ViLT-FT \cite{vilt} & 74.63 & 74.76& 75.21& 74.99\\
VLMO-FT \cite{vlmo} & 83.02 & 85.53& 81.94& 83.70\\
X-VLM-FT \cite{xvlm} & 84.16  & 87.01& 82.73& 84.82\\
\bottomrule
\end{tabular}
\caption{Results of various baselines on the NLVR2 dev set. We report all the numbers in \%}
\label{tab:results_table}
\end{table*}

\paragraph{Unimodal Baselines}
\subsection{Text-only baselines}
\textbf{RoBERTa} \citep{zhuang-etal-2021-robustly}: RoBERTa is a bidirectional language model that has been pre-trained using the Masked Language Modeling (MLM) objective. It has the same architecture as BERT~\citep{devlin-etal-2019-bert} but performs better than BERT on most NLP benchmarks due to a more robust fine-tuning strategy. We will use the RoBERTa-Large variant as our unimodal baseline.
\newline
\newline
\textbf{GPT} \citep{Radford2019LanguageMA, gpt}: GPT is an autoregressive language model pre-trained on the Causal Language Modeling task. GPT2 and GPT3 have achieved SOTA in many NLP benchmarks in the zero-shot and few-shot settings. We will use the open-source GPT-neo-1.3B model as a unimodal baseline which has a very similar architecture to GPT-2.
\newline
\newline
\textbf{T5:} \citep{raffel2020exploring} T5 is a text to text bidirectional transformer that was trained to predict sequences. It proposes to reframe all NLP tasks into a unified text-to-text-format where the input and output are always text strings. It was also trained using the MLM objective. We will be using Flan-T5~\cite{chung2022scaling} as a baseline that has been instruction finetuned and finetuned using Chain-of-Thought reasoning and performs much better than vanilla T5.

\subsection{Image-only baselines}
\textbf{ResNet:} \citep{resnet} We will use ResNet-50 as our unimodal image baselines. It is a Convolutional Neural Network (CNN) with skip connections to create a very deep architecture. 
\newline
\newline
\textbf{ViT:} \citep{vit} ViT or Vision Transformer is the popular transformer architecture applied to images. It was pre-trained using supervised learning on a large corpus of data. We will use the ViT-Large variant as our unimodal baseline.
\newline
\newline
\textbf{MAE:} \citep{mae} MAE or Masked Auto Encoders are Vision Transformers that have been pre-trained using the masked image-filling task. Given a randomly masked image, MAE's have been trained to reconstruct the original image. This is similar to the MLM objective for text. We will be using the MAE based on ViT-Large as our unimodal baseline. 

\subsection{Simple Multimodal Baselines}
\textbf{MAE+RoBERTa: } Our approach is straightforward: we employ two separate unimodal models to obtain embeddings, and then concatenate these embeddings before training a linear classifier on the resulting concatenated data. This is one of the obvious choices for a simple multimodal model. Using this model shows that pretraining on multimodal tasks is very important for doing well on this task which is further strengthened by the fact that ViLT(CLS) performs much better.

\textbf{ViLT-LP: }
The ViLT model is similar to a transformer, but its designed to handle multiple modes of data. It processes images using patches as tokens, instead of relying on convolutional architecture and image feature extraction. As a proof of concept for our proposed method where we plan to train a linear projection, we used a pre-trained ViLT model and trained a linear classifier on  top of it. Our goal was to see if a basic linear classifier on a fixed multimodal model could enhance performance on the NLVR2 dataset.

\subsection{Competitive Baselines}
We run $3$ competitive baselines available at public repos.  These include: \\
\textbf{ViLT-FT:} The ViLT-FT model involves training a linear classifier on the fixed pretrained ViLT backbone. We also benchmark ViLT-FT on the dev set where the complete ViLT backbone is finetuned on the train set. We use this model as ViLT is one of the most standard multimodal models and has a competitive performance on NLVR2.

\textbf{VLMO-FT:}
VLMO is a pre-training model that operates on multiple modalities and utilizes the concept of mixture of experts. Its goal is to understand the connections between images and text through contrastive training on a vast corpus of multimodal data. By employing contrastive training, VLMO is able to achieve a high level of alignment between different modalities, making it an ideal candidate for the NLVR2 task. Just like ViLT it also uses MLM and Image text-matching objectives for pretraining. We choose to use this model as it has a very competitive performance on NLVR2 and the authors release a finetuned checkpoint.

\textbf{X-VLM-FT: }
X-VLM uses a different approach compared to traditional methods, which typically use object detection techniques to locate image paths and then align text. Instead, X-VLM performs multi-alignment by identifying visual concepts based on the relevant text descriptions. Such visual concepts can be either objects, regions or also the entire image. We use X-VLM since like X-VLM we also propose to use object-level features in our proposed model and using this model helps us to validate our choice of focusing on object-level features. We again evaluate X-VLM in both settings FT and LP, similar to ViLT.

\begin{figure}
    \centering
    \includegraphics[width=8cm]{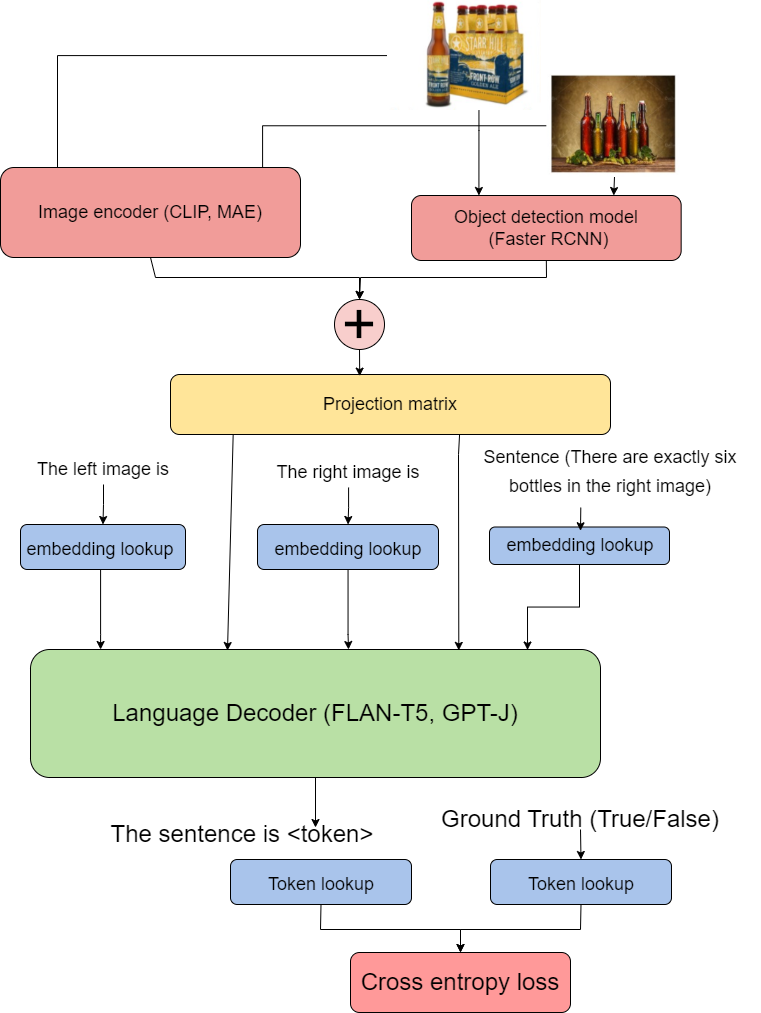}
    \caption{Model architecture}
    \label{model_architecture}
\end{figure}
\begin{table*}[ht]
\centering
\begin{tabular}{@{}lrrrr@{}}
\toprule
 & \multicolumn{2}{c}{FLAN-T5} & \multicolumn{2}{c}{OPT} \\
 \midrule
 & \multicolumn{1}{l}{Generation Loss} & \multicolumn{1}{l}{BCE Loss} & \multicolumn{1}{l}{Generation Loss} & \multicolumn{1}{l}{BCE Loss} \\
 \midrule
SAM + CLIP & 52.69 & 50.98 & 52.12 & 50.45 \\
SAM + MAE & 51.67 & 50.12 & 51.23 & 50.06\\ 
 \bottomrule
\end{tabular}
\caption{Accuracies obtained by using different loss metrics.}
\label{tab:diff_enc}
\end{table*}
\section{Proposed Model}
In this project, we provide a detailed analysis of various bridge architectures. A bridge architecture is a model that takes images as input, passes it via an image encoder, projects it onto the text space of an LLM via a linear projection, and then uses the zero-shot reasoning capabilities of the LLM to perform various tasks such as dialogue generation, question answering etc. The linear projection acts as a bridge between the vision encoder and the LLM. Figure~\ref{model_architecture} demonstrates our proposed bridge architecture. Given an image pair ($x_l$, $x_r$) and a sentence ($s$),
we initially pass both images through an image
encoder to get the global image features. We also
pass the image pair through a segmentation model like SAM~\cite{sam} to get the local image features
since NLVR2 requires us to have an object-level
understanding of the image. We concat both the
global and the local features and then pass them
through the bridge to convert
them as inputs to the LLM. We also add
the sentence as input and then pass them through
the LLM to get our predictions. The only
trainable part of the whole pipeline is the image to
text projection matrix making our pipeline very fast
and efficient to train. We tried using CLIP and MAE and our image encoders, SAM as the segmentation model and FLAN-T5 and OPT as our LLMs. We provide results of using different image encoder and LLMs in Table~\ref{tab:diff_enc}. We notice that using the instruction finetuned FLAN-T5 gives slightly better results than OPT although the improvement is not significant. This is expected as we see that our further analysis indicates that instruction finetuning helps. In addition to this we also include three off-the-shelf pretrained bridge architectures namely \textbf{Fromage}~\cite{fromage}, and \textbf{LLaVA}~\cite{llava}, \textbf{Open Flamingo} \cite{open-flamingo}. LLaVA is a bridge architecture that has been pretrained on cc3m dataset on the captioning task and finetuned on multimodal instruction data. Fromage is a bridge architecture that has been pretrained on cc3m data on the captioning as well as retrieval task. Open-Flamingo is an architecture that projects the image into the LLM space via a cross-attention layer into every layer of the LLM and pre-trained on the LAOIN-2B~\cite{laion} dataset. In the case of LLaVA due to the unavailability of A100 GPUs we evaluate it in the zero-shot setting by trying to leverage it's chain of thought capabilities and using bootstraping to improve it's performance by feeding it's own chain of thought. We also evaluate Flamingo in the zero-shot setting. In the case of Fromage, we use the embeddings of the [RET] token to finetune a linear classifier layer using the standard BCE loss. This choice of embeddings make sense since in Fromage the embeddings of the [RET] token are computed by attending over both the project image and the text embeddings and thus have multimodal information encoded. 
\subsection{Loss functions}
We try out two different loss functions for our model. We present the results of using different loss functions in Table~\ref{tab:diff_enc}.
\paragraph{Binary CrossEntropy Loss: }Given a sentence $s$, we append it with "This statement is" and apply BCE loss corresponding to the predicted logits of "True" and "False" by model for the next word.
\paragraph{Generation Loss: }Given a sentence $s$, we append it with "This statement is True/False" depending on the label and then we use the standard cross entropy loss used for auto-regressive language modeling in the case of OPT and seq2seq language modeling in the case of FLAN-T5.
We notice that using Generation loss is slightly more beneficial than using BCE loss although it is not significantly better. This is in alignment with the findings by the authors of \citet{sachin} where they demonstrate that finetuning vision models in sync with the pretraining regime are better.
\subsection{Changes to training data}
We don't make any changes to the training data. During zero-shot inference, since LLaVA is designed to take a single image, we fuse the two images and give them as input to LLaVA. We also provide a prompt an LLaVA to elicit a chain of thought prompting and improve its zero-shot performance.
\subsection{Hyperparameters and their effects}
\begin{figure}[H]
    \centering
    \includegraphics[width=0.5\textwidth]{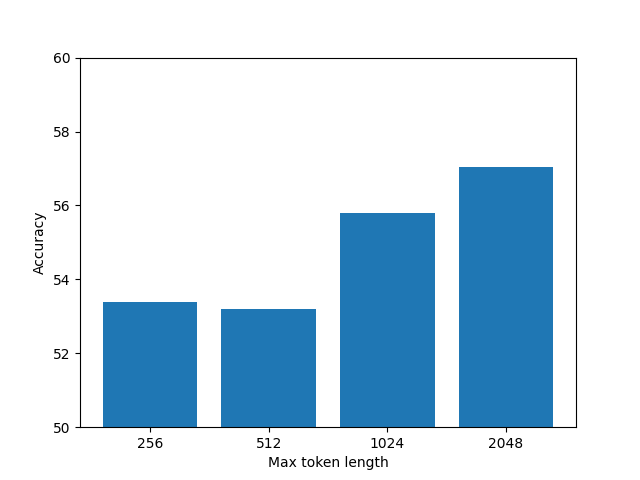}
    \caption{Variation of Accuracy in Chain of though Prompting with max token length. Having a large token length allows the model to reason better and hence choose the answer more accurately.}
    \label{fig:acc}
\end{figure}

We performed extensive tuning of train batch size, learning rate and number of visual tokens for training our model (SAM + FLAN + LIMBER) but unfortunately, the best we could reach was a performance of $52.69\%$
Showin in Figure \ref{fig:acc} is the variation of accuracy of LLaVA with max token length. Using a larger token length allows the model to build stronger reasoning about the images and hence choose the answer more accurately.

\section{Analysis }
\begin{figure*}[ht]
    \centering
    \begin{subfigure}[b]{0.3\linewidth}
        \includegraphics[width=\linewidth]{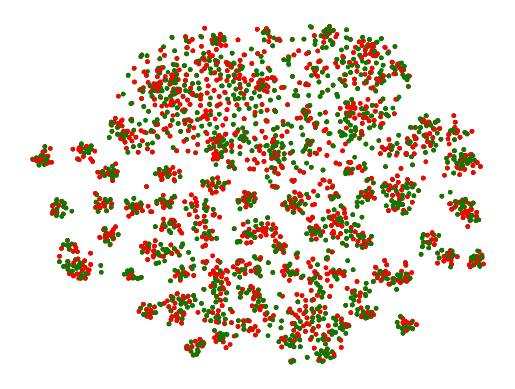}
        \caption{Fromage embeddings by label}
        \label{fig:image1}
    \end{subfigure}
    \hfill
    \begin{subfigure}[b]{0.3\linewidth}
        \includegraphics[width=\linewidth]{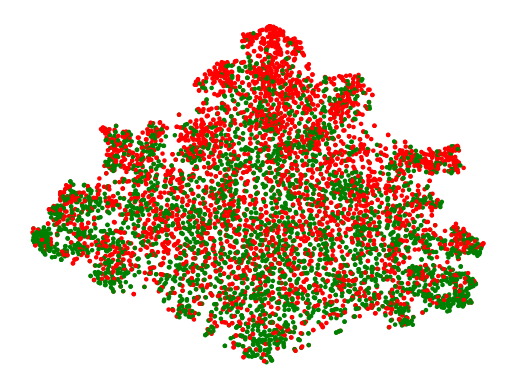}
        \caption{X-VLM embeddings by label}
        \label{fig:image2}
    \end{subfigure}
    \hfill
    \begin{subfigure}[b]{0.3\linewidth}
        \includegraphics[width=\linewidth]{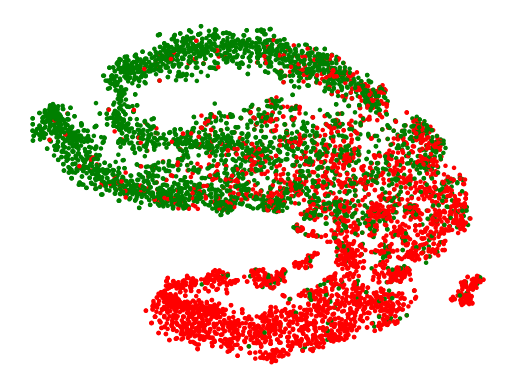}
        \caption{X-VLM FT embeddings by label}
        \label{fig:image3}
    \end{subfigure}
    
    \medskip
    
    \begin{subfigure}[b]{0.3\linewidth}
        \includegraphics[width=\linewidth]{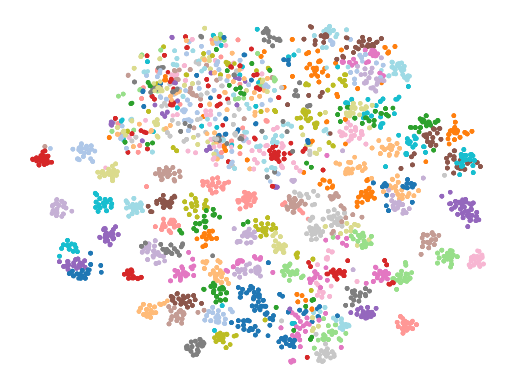}
        \caption{Fromage embeddings by synset}
        \label{fig:image4}
    \end{subfigure}
    \hfill
    \begin{subfigure}[b]{0.3\linewidth}
        \includegraphics[width=\linewidth]{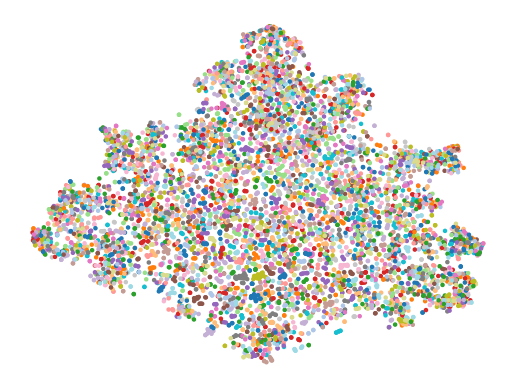}
        \caption{X-VLM embeddings by synset}
        \label{fig:image5}
    \end{subfigure}
    \hfill
    \begin{subfigure}[b]{0.3\linewidth}
        \includegraphics[width=\linewidth]{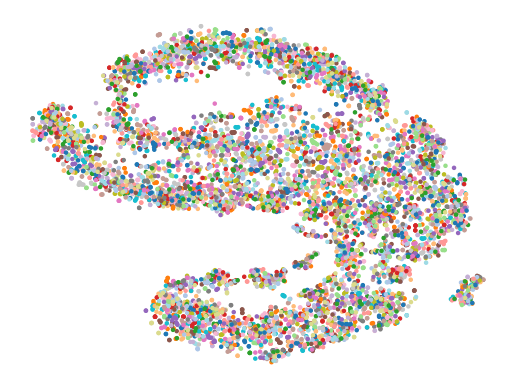}
        \caption{X-VLM FT embeddings by synset}
        \label{fig:image6}
    \end{subfigure}
    \caption{Caption for the entire figure}
    \label{fig:tsne}
\end{figure*}

We summarize the models that we considered in Table \ref{tab:models}. \textbf{FT} stands for finetuning i.e  the model was finetuned on NLVR2 and \textbf{LP} (Linear Probing) implies that we finetuned only a linear classifier layer on the embeddings of the pretrained model. We also evaluate the zero-shot capabilities of LLaVA on NLVR2. We provided the prompt "Given two images side by side predict if the following statement is True or False in a single word". This proved to be ineffective and the model accuracy was equivalent to that of random guessing. \cite{cot} showed that using chain of thought prompting elicits reasoning in large language models. Since LlaVA was multimodally instruction finetuned we leverage the same for our model (LLaVA ZS + COT). CoT provided a small improvement in performance which demonstrates the dormant zero-shot multimodal capabilities of LLaVA. Next, we tried to bootstrap LLaVA by feeding it its own CoT reasoning (LLAVA ZS + COT + BS) in the same vein as \cite{selfimprove}. But unlike ~\citet{selfimprove}, we dont finetune LLaVA using bootstrapping but evaluate it in the zero-shot setting. Although the performance is still worse than linear probe results of X-VLM and ViLT, our bootstrapping method provides a small boost in performance which demonstrates the zero-shot capabilities of LLaVA and gives us hope for bridge architectures. We provide some qualitative examples in Table~\ref{tab:cot_table} where we can see that LLaVA with CoT is able to reason out the answer and we also see an example where LLaVA with CoT bootstrapping is able to correct its own answer. We believe LLaVA would perform better if we make it reason step by step in conversational mode as done by~\citet{cot}. This is an aspect we will explore in our future work. \\
The first half of the table summarizes state-of-the-art multimodal models, we observe that these models perform well on the NLVR2 task as can be seen from Table \ref{tab:models}. Notably, even with just a linear probe we achieve significant improvements in performance. \\
The second half summarizes bridge architectures that we consider. The main reason we explore these architectures is that they don't require expensive finetuning of the image encoder on NLVR2. These architectures lack explicit crossmodal interactions unlike models like X-VLM which learn a joint image-text embedding. All crossmodal interactions occur in the language model when self-attention is applied over the sequence of projected image tokens and language tokens. The relatively strong zero-shot results of LLaVA demonstrate that multimodal pretraining and multimodal instruction finetuning is pretty important for bridge architectures. We posit that the massive difference in the performance of LLaVA over SoTA baselines is due to the image encoder being frozen during the pretraining and instruction finetuning stage which makes the multimodal interactions weak. We look to test our hypothesis in the near future.\\
We observe that when we only train a projection layer between the image and language space the model doesn't perform well on the NLVR2 dataset. Our hypothesis is that NLVR2 requires specific types of finegrained visual information that the image encoder is not trained to retain. To demonstrate some of these points we generate t-SNE plots of the fused visuo-lingual representations from some models, see Figure \ref{fig:tsne}\\ 
The first and third plots compare fromage and X-VLM finetuned embeddings by label (True/False). X-VLM embeddings are well separated, however when the same points are colored by synset (subject of the image) labels we observe almost no well formed clusters. This seems to indicate that synset level information is not necessary to perform well on the NLVR2 task. However, Fromage \cite{fromage} having been trained on captioning and retrieval task that requires strong object understanding to perform well on these tasks hence we observe good synset-wise cluster separation in the t-sne plot. By comparing the second and third plots we observe that finetuning on NLVR2 generates embeddings that are well separated. This points to the importance of finetuning for good performance on NLVR2. X-VLM embeddings regardless of finetuning are not synset separable.\\

\begin{table*}[t]
\footnotesize	
\scalebox{0.9}{
\begin{tabular}{@{}lcccccc@{}}
\toprule
\multirow{2}{*}{}   & Multimodal Pretraining & \multicolumn{2}{c}{Instruction finetuning} & Image encoder frozen & Crossmodal interactions & Accuracy\\
\midrule
                             &           & Unimodal            & Multimodal           &               &                           \\
                    \midrule
ViLT-LP  & \CheckmarkBold             & &  &  \CheckmarkBold   & \CheckmarkBold  & 62.10 \\
ViLT-FT      &   \CheckmarkBold               &&    &&   \CheckmarkBold &  74.63\\
X-VLM-LP &  \CheckmarkBold              & &   & \CheckmarkBold & \CheckmarkBold & 69.90    \\
X-VLM-FT   &   \CheckmarkBold            & &   &     & \CheckmarkBold & \textbf{84.16}\\
 \midrule
 Open Flamingo & \CheckmarkBold & &&\CheckmarkBold& \CheckmarkBold   & 51.06\\
Fromage-LP          &   \CheckmarkBold            & \CheckmarkBold&    &  \CheckmarkBold & &51.10    \\
SAM+FLAN+LIMBER     &                &\CheckmarkBold &    &   \CheckmarkBold & & 52.69 \\
LLAVA ZS &     \CheckmarkBold       &         &    \CheckmarkBold & \CheckmarkBold & & 52.74
\\
LLAVA ZS + COT     &  \CheckmarkBold         &         &    \CheckmarkBold & \CheckmarkBold & & 54.06
\\
LLAVA ZS + COT + BS &    \CheckmarkBold         &         &    \CheckmarkBold & \CheckmarkBold & & \textbf{57.04}
\\
\bottomrule\\
\end{tabular}
}
\caption{Performance of various multi-modal models under different settings. (Accuracy is in percentage)}
\label{tab:models}
\end{table*}

\begin{table*}[ht]
\centering
\begin{tabular}{|c|c|p{4cm}|p{6cm}|}
\hline
\textbf{Left Image} & \textbf{Right Image} & \textbf{Input Statement} & \textbf{Comments} \\
\includegraphics[width=3cm]{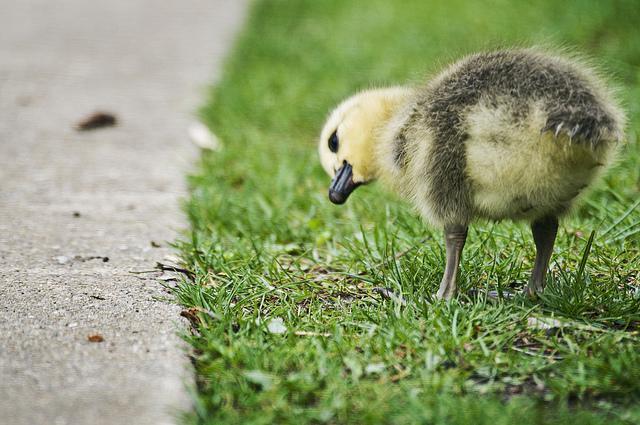} & \includegraphics[width=3cm]{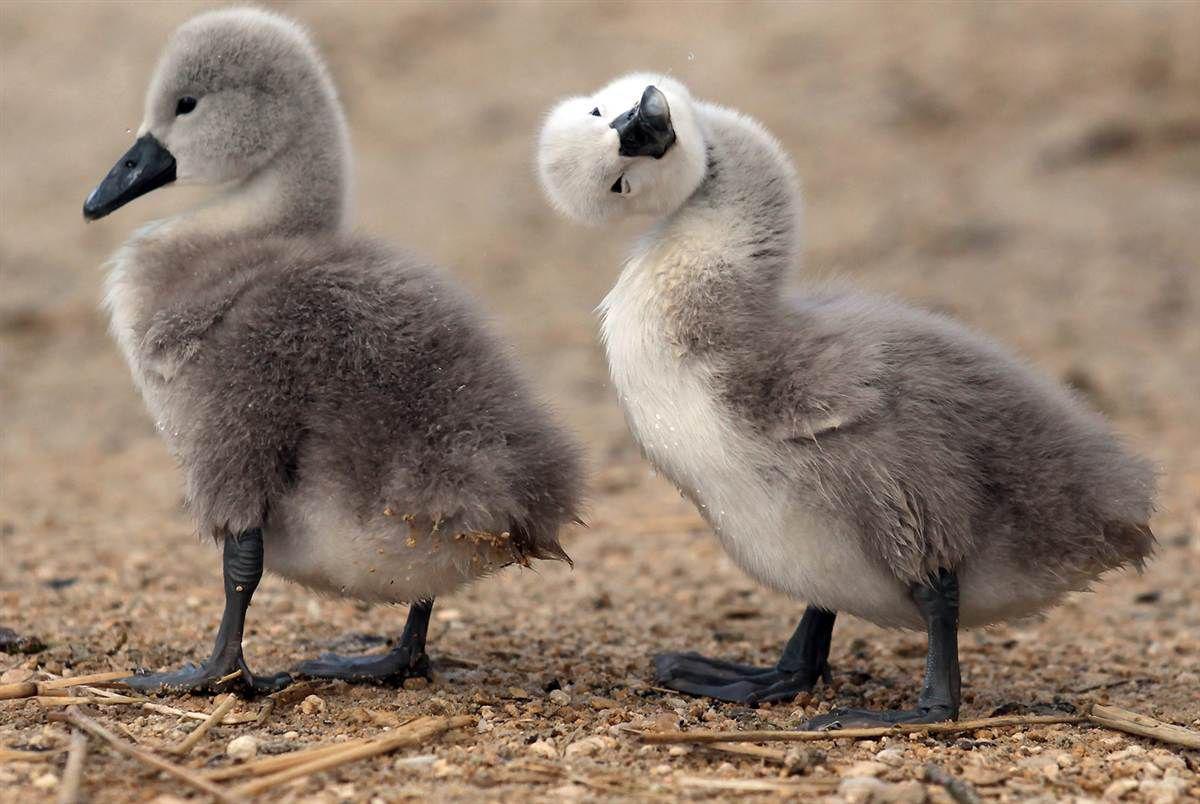} & \vspace{-2cm}\textbf{Statement:} There are exactly two ducks in the right image. &\vspace{-2cm} \textbf{LlaVA ZS}: False \;\;\;\;\;\qquad\quad\quad\quad \textbf{LLaVA ZS+COT}:  In the left image, there is a duckling standing in the grass, while in the right image, there are two ducks standing in the grass. Hence, the statement is True\\
\hline
\includegraphics[width=3cm]{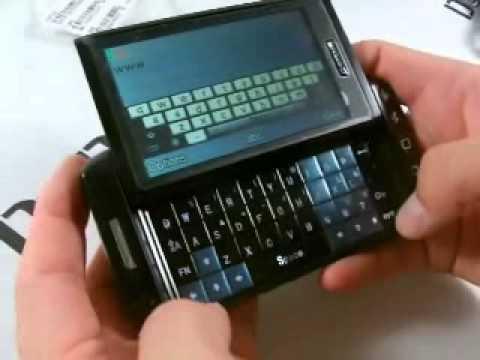} & \includegraphics[width=3cm]{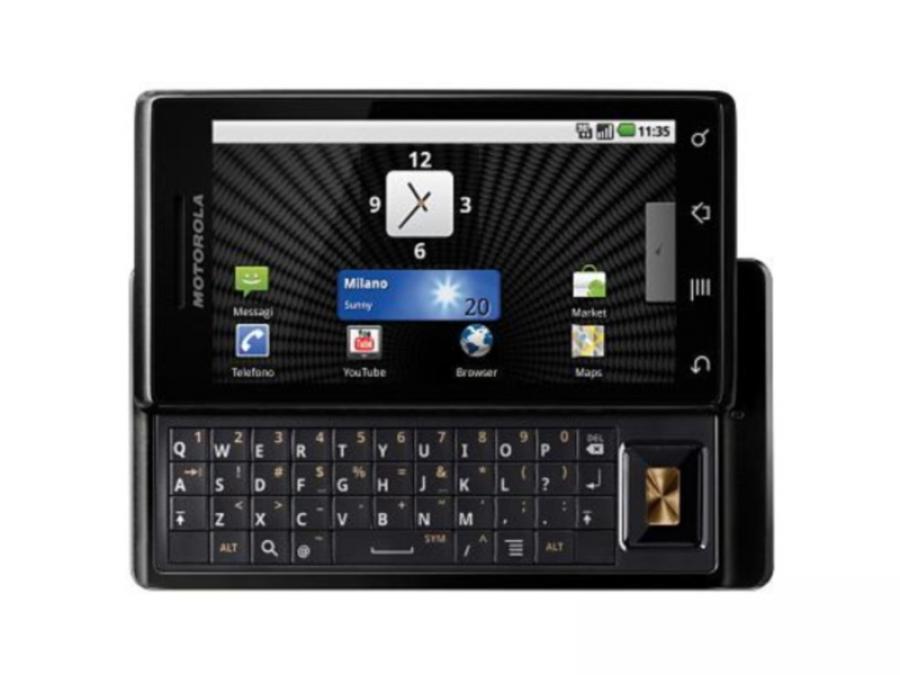}  & \vspace{-2cm}\textbf{Statement:} The keyboard of the phone in one of the images is pink. \qquad\qquad  \quad\textbf{Reasoning (LLaVA ZS+COT):} The statement to be reasoned out is: "The keyboard of the phone in one of the images is pink."Using the images, we can observe that there are two cell phones. One of them has a keyboard that is visible and seems to have a pink key. We can conclude that the statement is True.& 
\vspace{-2cm}
\textbf{LLaVA ZS+COT+BS}: The statement is false. In the image, there is a person holding a cell phone with its keyboard covered, and the keyboard is not pink. The image shows a black cell phone with a black keyboard, and there is no visible pink keyboard.\\

\hline
\end{tabular}
\caption{Examples where LLaVA performs decent reasoning}
\label{tab:cot_table}
\end{table*}

\subsection{Qualitative Analysis and Examples }

We present the qualitative failure cases for all the models in Tables \ref{tab:uni_qual_res}, \ref{tab:smul_qual_res}, \ref{tab:cmul_qual_res} and \ref{tab:bridge_fail}. 
\subsubsection{Uni-modal Failure Cases: } As we can see in Table \ref{tab:uni_qual_res}, for the uni-modal baselines, they fail even on very simple examples. This is expected since all these models only take into account one of the modalities while making decisions. For example, in Row 2 for Table \ref{tab:uni_qual_res}, the example on gorrilas could probably be solved by very simply multi-modal baselines which simply keep track of the object count and can do basic semantic parsing. However, the text only model (RoBERTa) fails on this. Similarly in row 4, the image only model (MAE) fails on a relatively simple example which would've been correct if there were some text signals present. Since only one modality was involved in making inferences, we cannot analyze the failure cases further in depth.
\subsubsection{Simple multi-modal Failure Cases: } As it is expected, these failure cases become harder than the previous ones, because these models take into account both modalities while making inferences. As it can be seen in the first row of Table \ref{tab:smul_qual_res}, the MAE+RoBERTa model most proabably fails because we only use the global feature vector from MAE, and it lacks object level local features, which are required to answer statements such as comparing the count of objects between the two images. The ViLT (CLS) model fails on an example in row 5 of Table \ref{tab:smul_qual_res}, which we can attribute to the lack of dense cross-modal interactions, since ViLT would probably see the words crabs and water and predict that crabs would be in the water (due to the abundance of such training samples in the pre-training step), whearas here the crabs are actually out of water. 
\subsubsection{Competitive multi-modal Failure Cases: } We can see that the examples in Table \ref{tab:cmul_qual_res} are extremely challenging to get correct. They require very fine-grained and dense cross-modal understanding of the sentence and the object and global features in the two images. For example, in row 3 in table \ref{tab:cmul_qual_res}, just the text query would probably be very difficult to understand fully for even the SOTA langauge models, since the sentence has a lot of details such as "steep sloped front", "black accent colors", and "angled rightward". Moreover, locating these features and distinguishing between the left and right image is even harder, for X-VLM as well. We see that row 1 of Table \ref{tab:cmul_qual_res} fails, perhaps because VLMO was not trained with object level features, but rather just image-text matching and contrastive learning. Thus it would be hard for the model to exactly get the count of bananas in the image. These results imply that object level features are extremely important for most models to perform well, because a lot of the questions have fine-grained information about object counts. Thus we plan to incorporate these local object-level features in our models. \
\subsubsection{Bridge Architecture Failure Cases:}
We report the failure cases of different bridge architectures in Table~\ref{tab:bridge_fail}. As is the case with almost all examples in this dataset, these are very challenging for a model and require multi-layered reasoning. From the t-sne plots, we can notice that Fromage is quite good at detecting the objects in the images but it lacks the complex reasoning ability required to reason over such examples and it is quite evident in the failure mode we have provided in Table~\ref{tab:bridge_fail} which requires reasoning over the position of the crabs and also the color. Detecting whether it is a crab or not will not be helpful because this question is object agnostic. Now we notice that Open-Flamingo fails in an example where LLaVA succeeds. That example requires complex reasoning and we see how the LLM in LLaVA due to its visual instruction finetuning is able to reason in Table~\ref{tab:cot_table} but Open-Flamingo fails on it inspite of using the same LLM (LLaMA). Now coming to the failure cases of LLaVA, we see that the model gets confused in the left and right images. We believe this happens because we fuse the two input images into one while providing it's input. We look forward to addressing this issue as a future direction. The failure case of SAM+FLAN+LIMBER is also illustrated in the Table. Since we freeze both the vision encoder and the language encoder when finetuning on NLVR2, our model is unable to retain complex visual features needed for the task. For instance to get this example right the model has to understand what a "young zebra" and "grazing" are. The vision encoder also has to preserve the directional information of the objects and only if all these are correct, the model makes an accurate prediction.

\subsubsection{Comparison of Competitive multi-modal Failure Cases: } In Table \ref{tab:cmul_comp}, we compare the three competitive multi-modal baselines on failure cases. We see that in row 1, all three models fail, which could perhaps be because of the difficult correlation of "unfolded towel" and the corresponding image, because in most training examples for image text matching or contrastive learning, such examples which include an unfolded towel are quite rare. Overall, we see that X-VLM performs better than VLMO and ViLT on tougher examples, which is consistent with our quantitative evaluations and analysis. 
\begin{table*}[ht]
\centering
\begin{tabular}{|c|c|p{5cm}|p{5cm}|}
\hline
\textbf{Left Image} & \textbf{Right Image} & \textbf{Text} & \textbf{Comments} \\
\hline
\includegraphics[width=3cm]{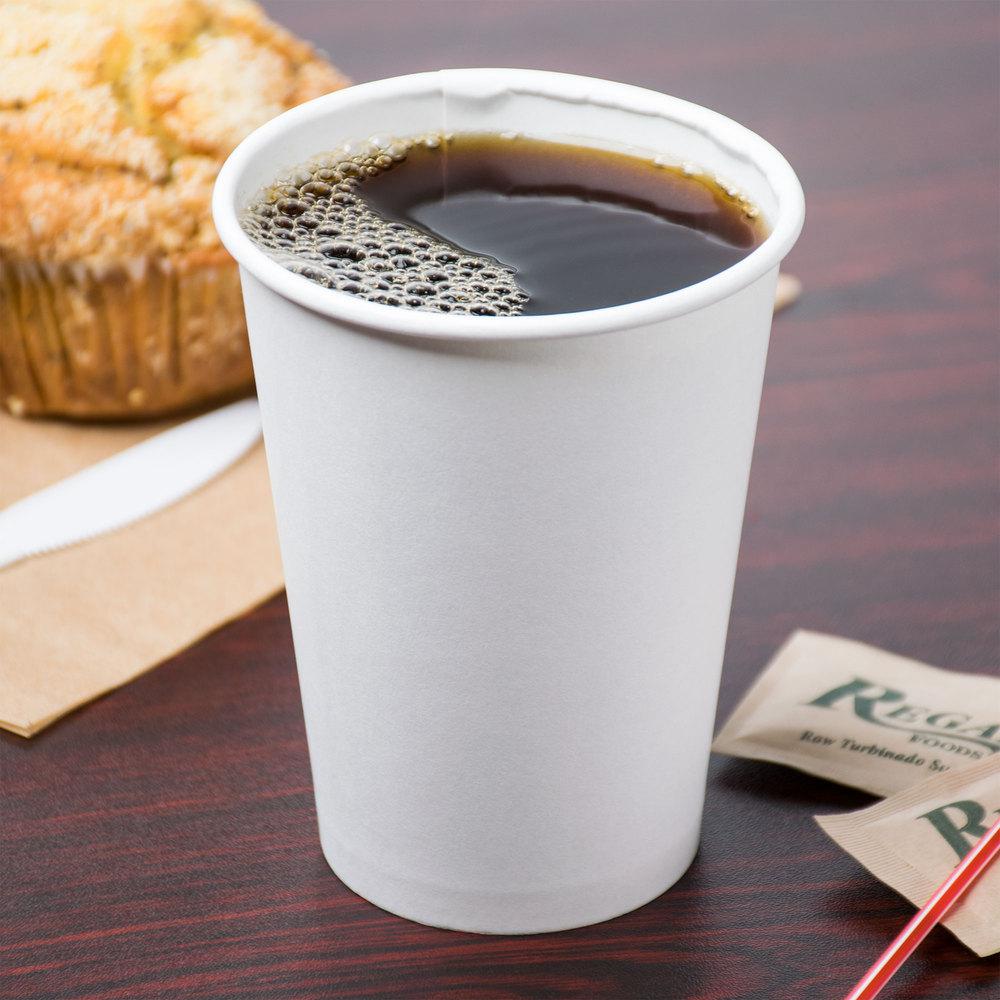} & \includegraphics[width=3cm]{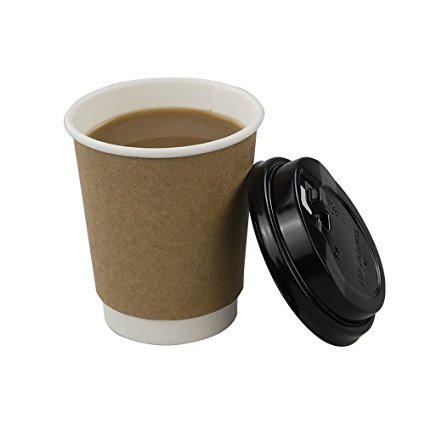} & At least one cup sits among several spread out coffee beans. & Failure Case for RoBERTa Model, Model Prediction: False \\
\hline
\includegraphics[width=3cm]{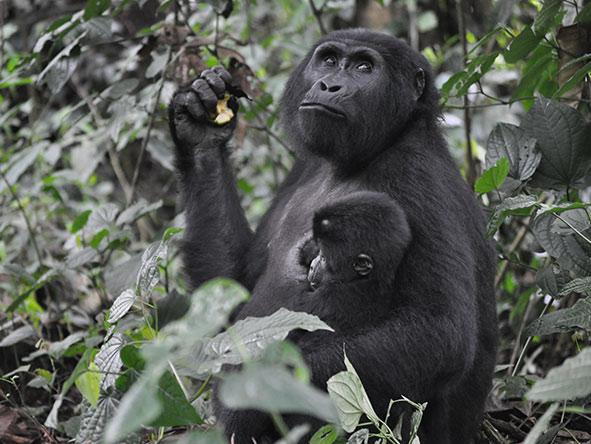} & \includegraphics[width=3cm]{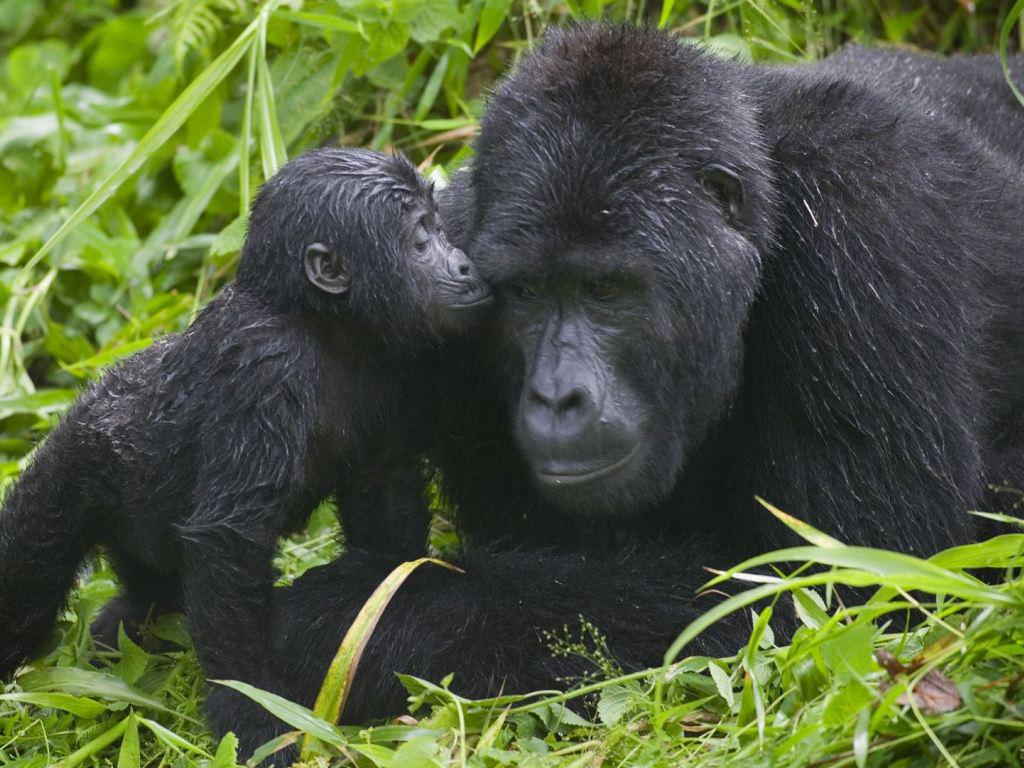} & There is exactly one gorilla in the right image. & Failure Case for RoBERTa Model, Model Prediction: False\\
\hline
\includegraphics[width=3cm]{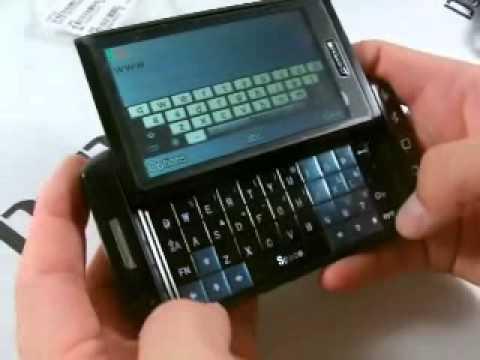} & \includegraphics[width=3cm]{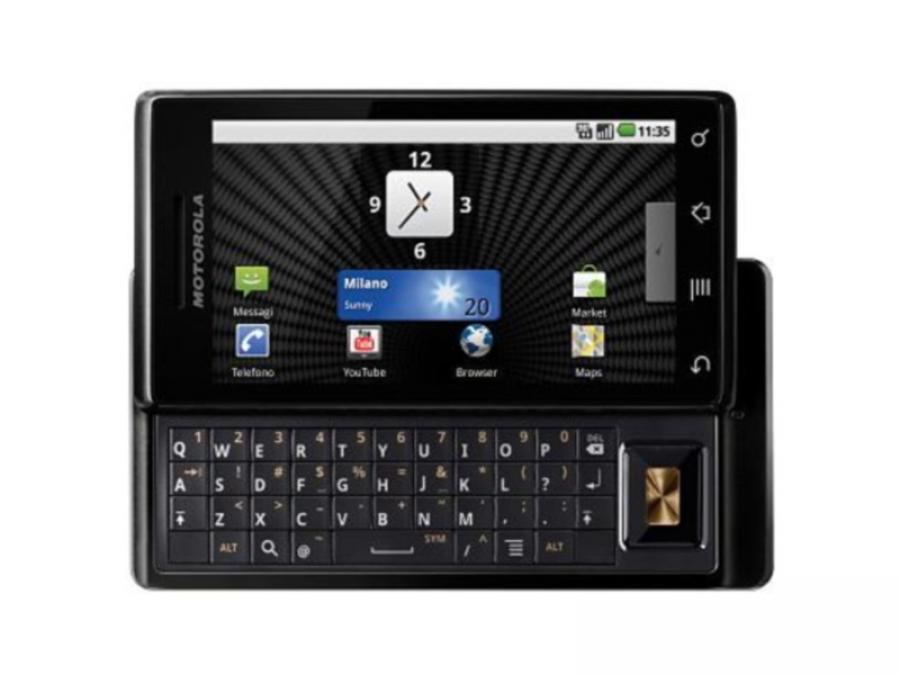} & The device on the right is viewed head-on, while the one on the left is angled facing rightward, and both devices have the screen slid up to show the keyboard. & Failure Case for MAE Model, Model Prediction: False\\
\hline
\includegraphics[width=3cm]{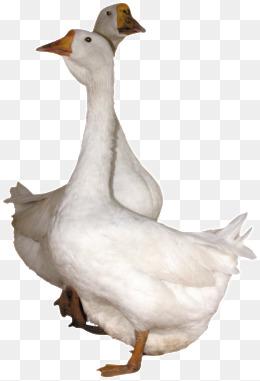} & \includegraphics[width=3cm]{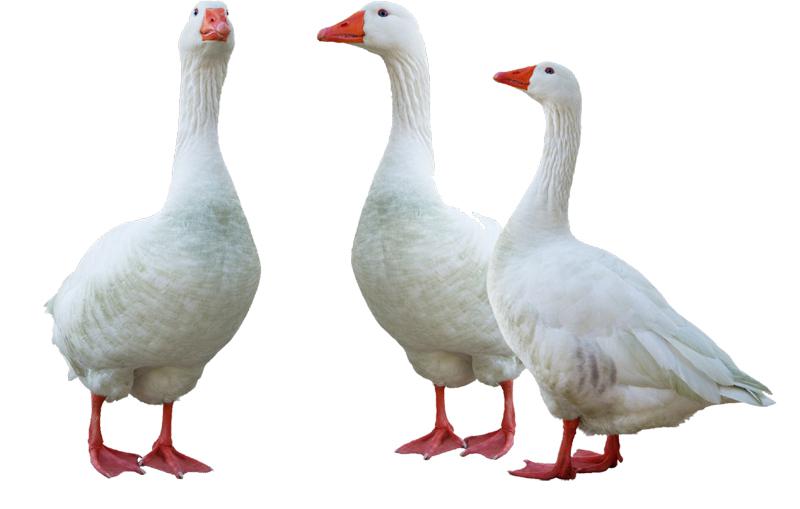} & Two birds are flying in the air. & Failure Case for MAE Model, Model Prediction: True\\
\hline
\includegraphics[width=3cm]{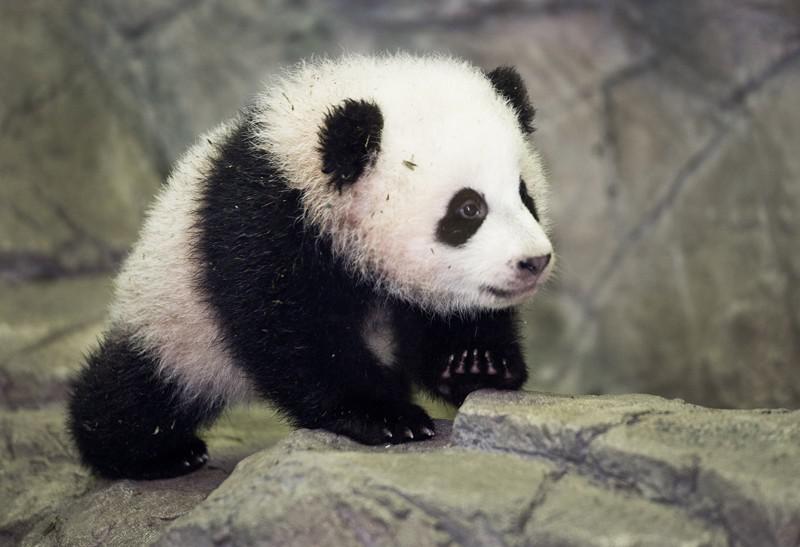} & \includegraphics[width=3cm]{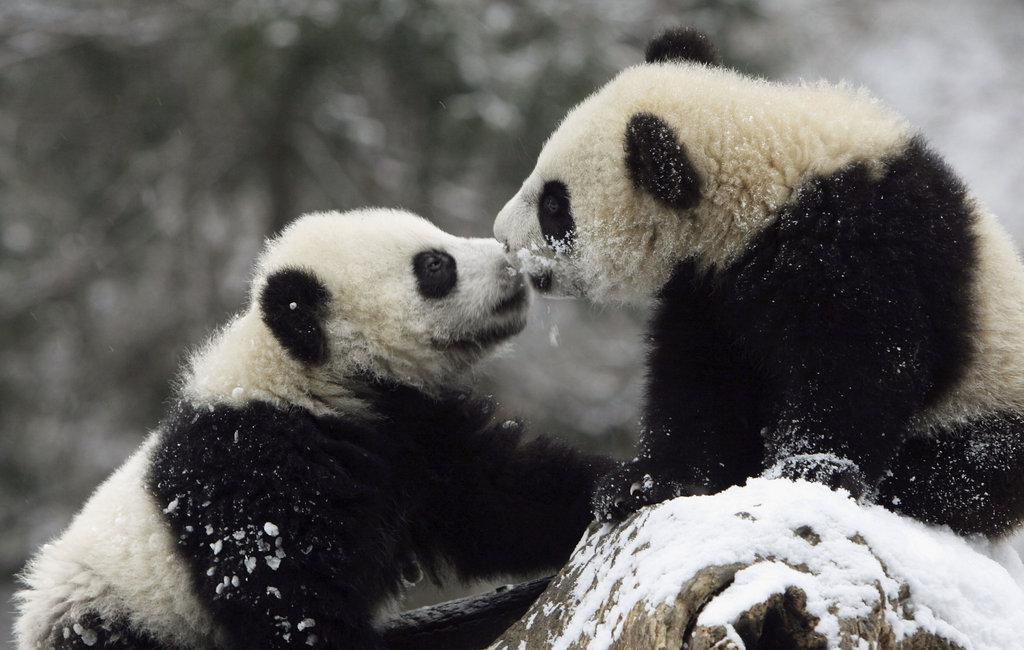} & Two pandas are touching mouths and one of them has at least one paw on the other panda's face. & Failure Case for ResNet-50 Model, Model Prediction: True\\
\hline
\includegraphics[width=3cm]{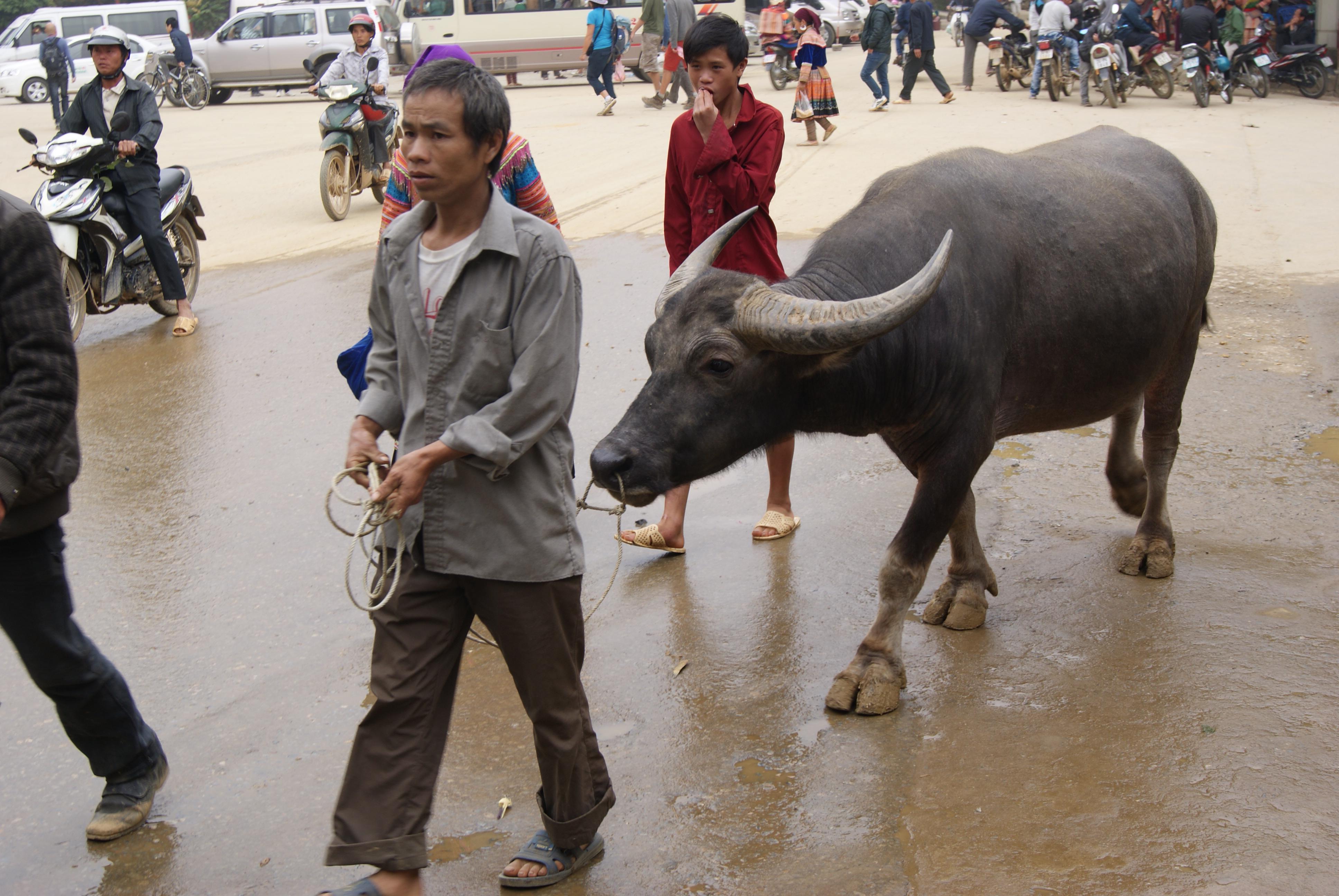} & \includegraphics[width=3cm]{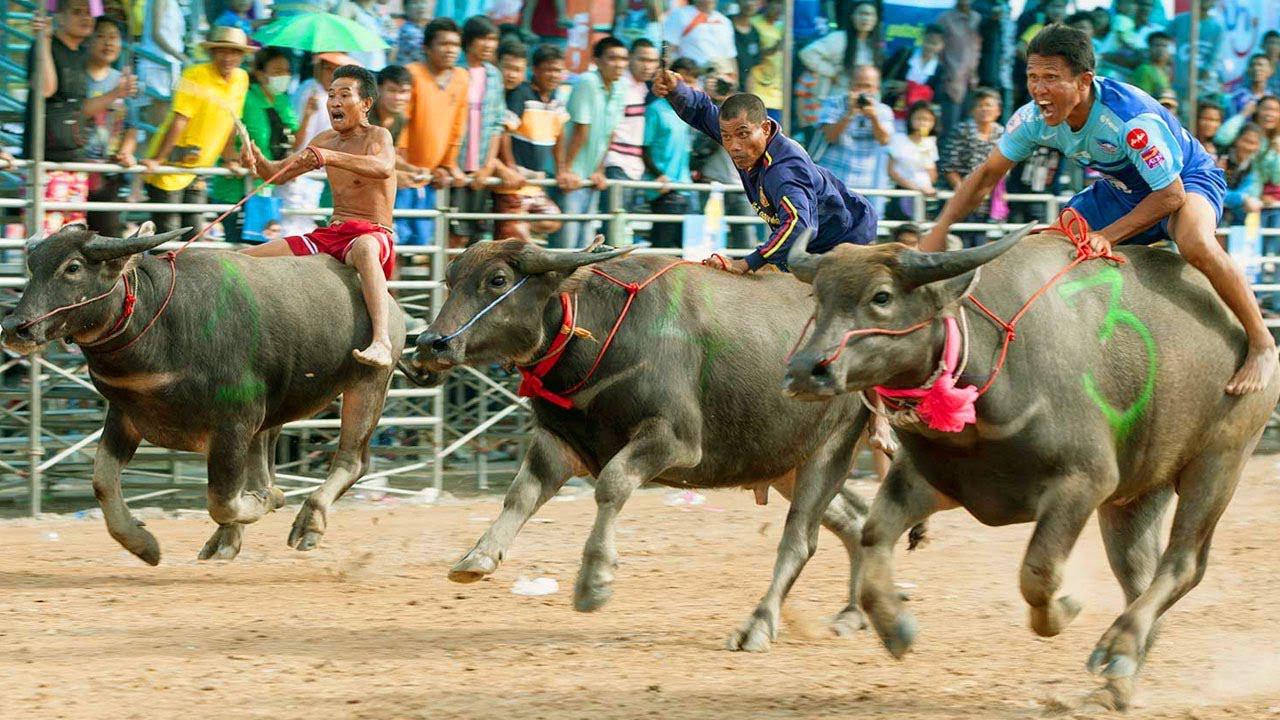} & The right image contains no more than one water buffalo. & Failure Case for ResNet-50 Model, Model Prediction: True\\
\hline

\end{tabular}
\caption{Failure Cases for Unimodal Models}
\label{tab:uni_qual_res}
\end{table*}

\begin{table*}[ht]
\centering
\begin{tabular}{|c|c|p{5cm}|p{5cm}|}
\hline
\textbf{Left Image} & \textbf{Right Image} & \textbf{Text} & \textbf{Comments} \\
\hline
\includegraphics[width=3cm]{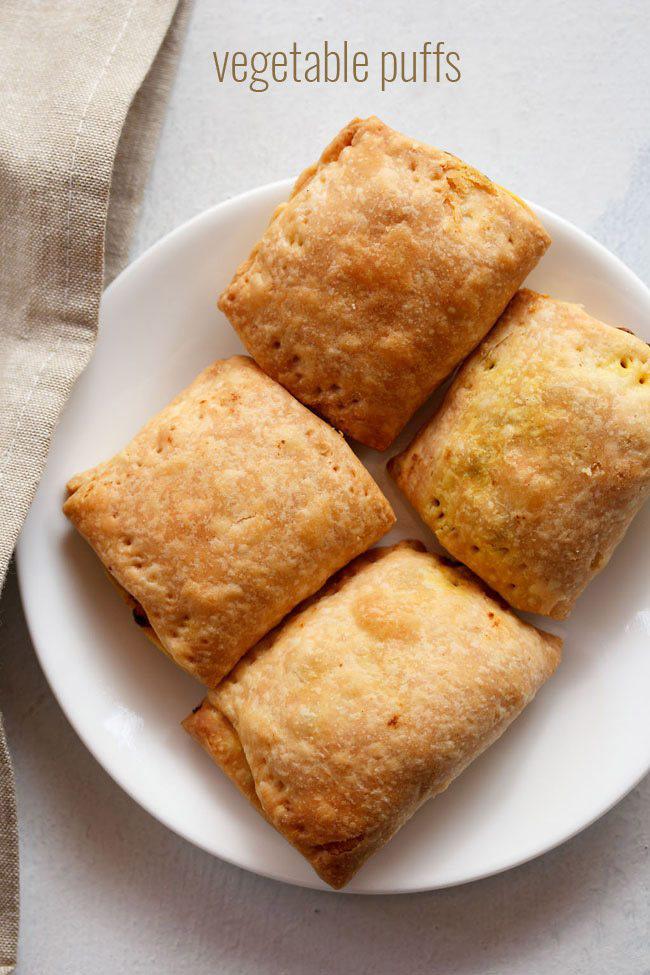} & \includegraphics[width=3cm]{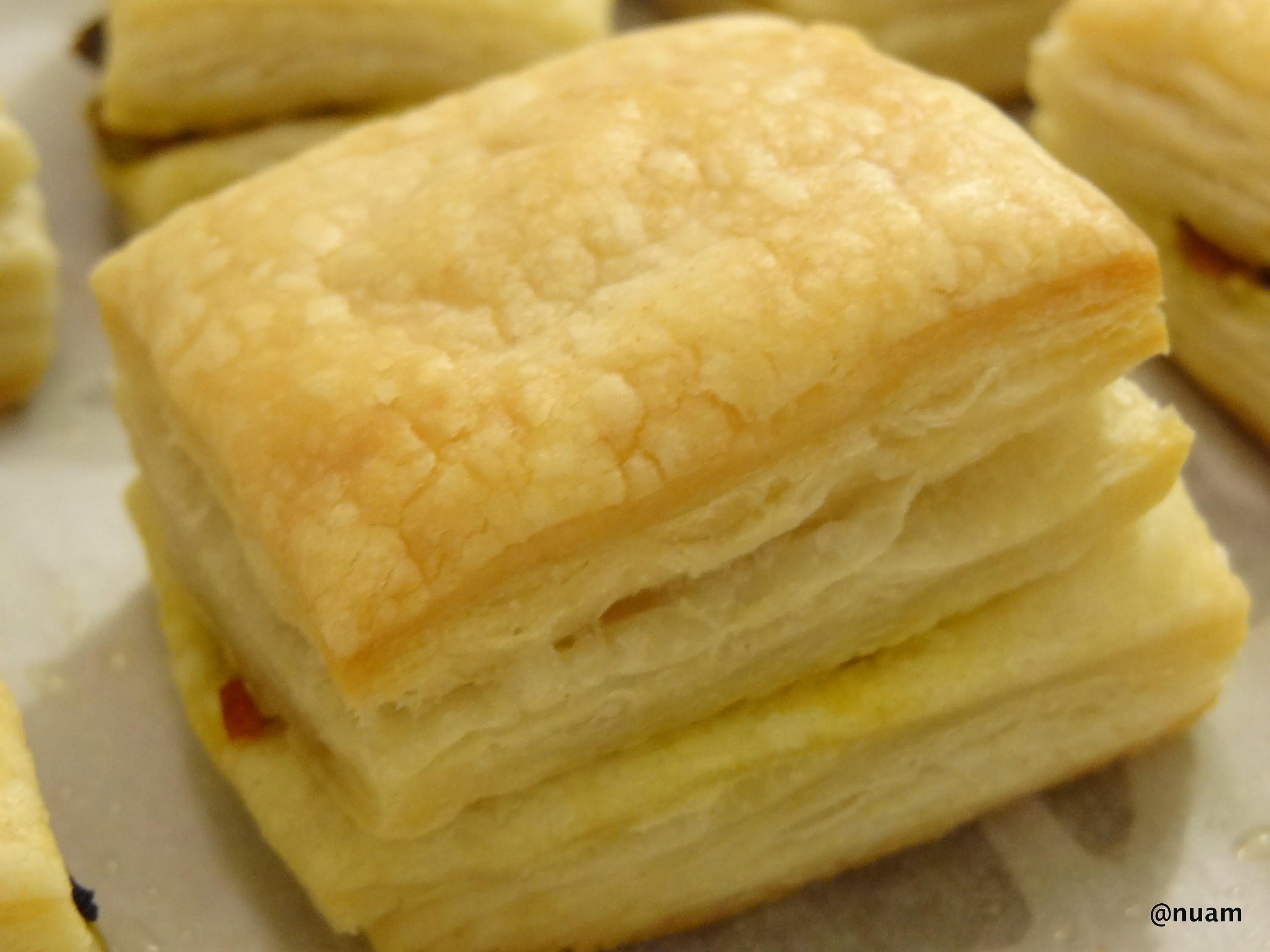} & There are at least three cooked triangle calzones displayed. & Failure Case for MAE+RoBERTa Model, Model Prediction: True\\
\hline
\includegraphics[width=3cm]{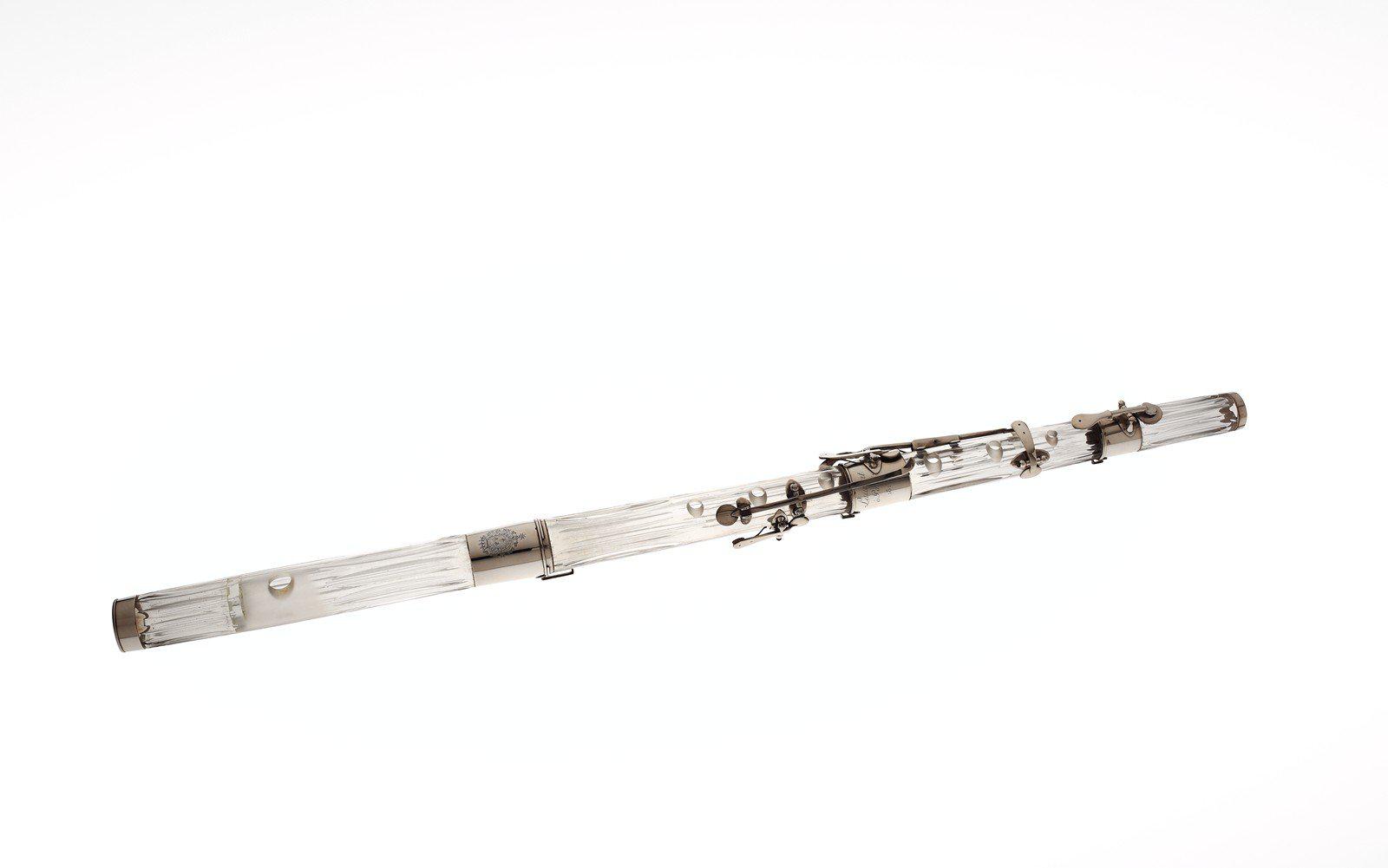} & \includegraphics[width=3cm]{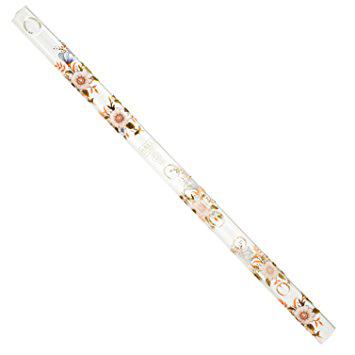} & One flute is resting on a holder. & Failure Case for MAE+RoBERTa Model, Model Prediction: True\\
\hline
\includegraphics[width=3cm]{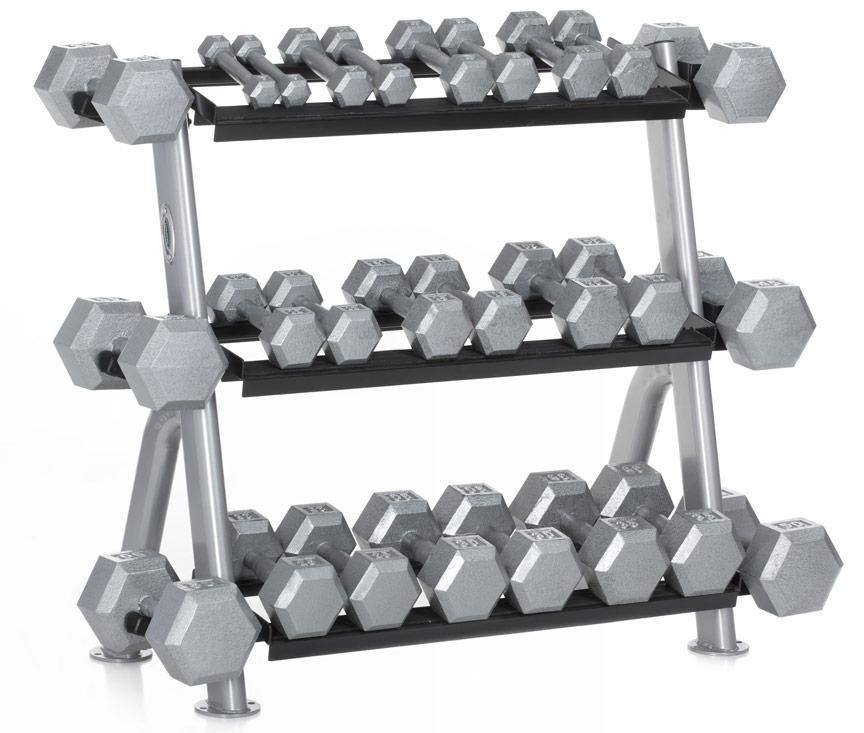} & \includegraphics[width=3cm]{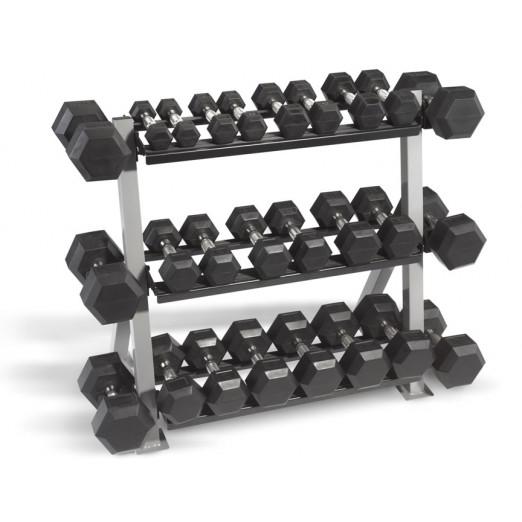} & There are two weight racks full of black weights. & Failure Case for ViLT (CLS) Model, Model Prediction: True\\
\hline
\includegraphics[width=3cm]{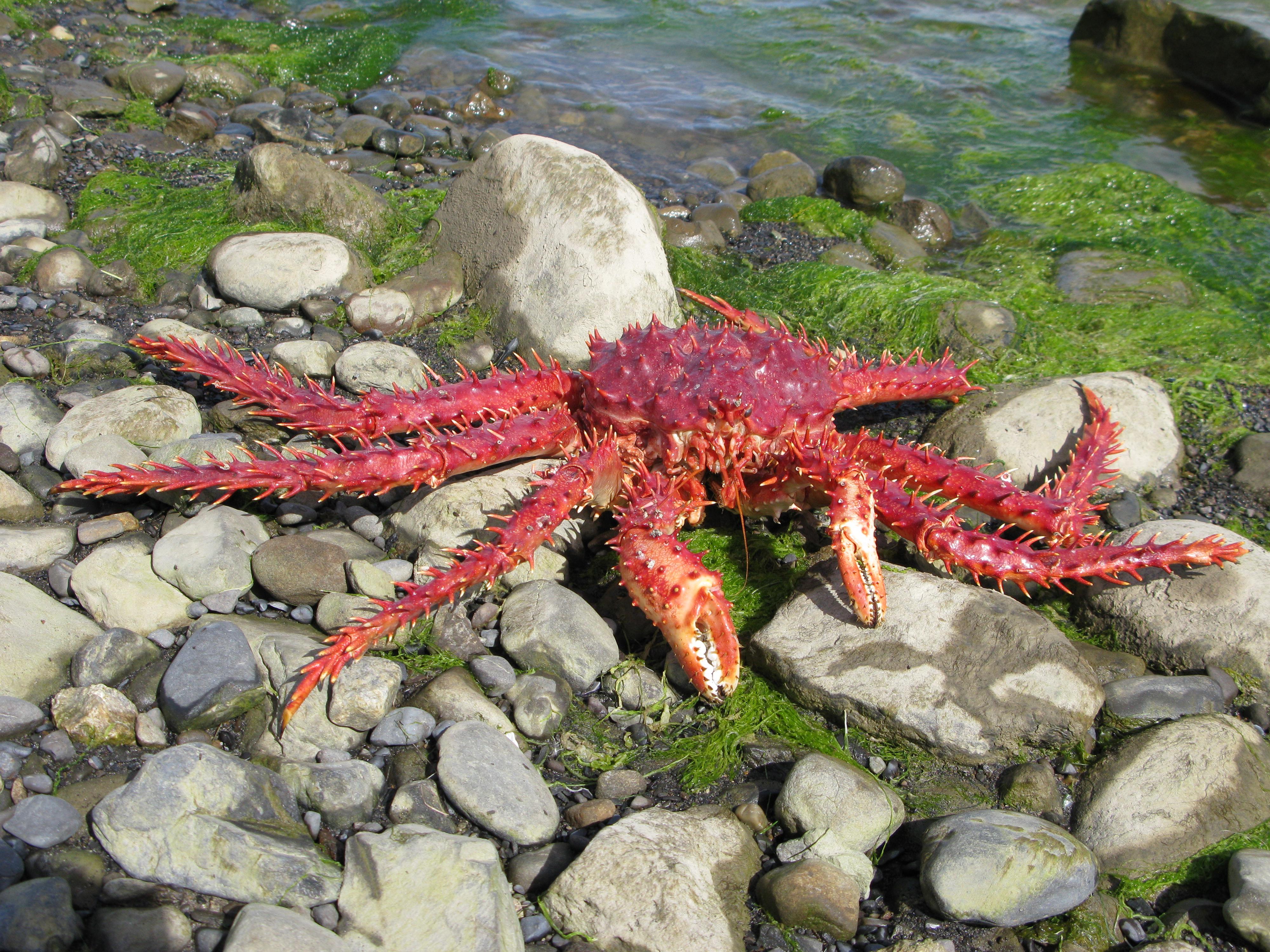} & \includegraphics[width=3cm]{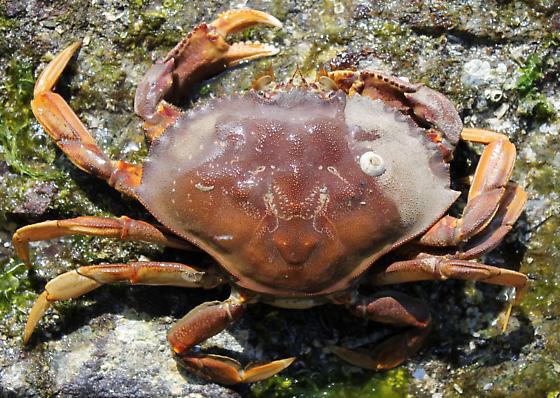} & All the crabs are out of the water. & Failure Case for ViLT (CLS) Model, Model Prediction: False\\
\hline
\includegraphics[width=3cm]{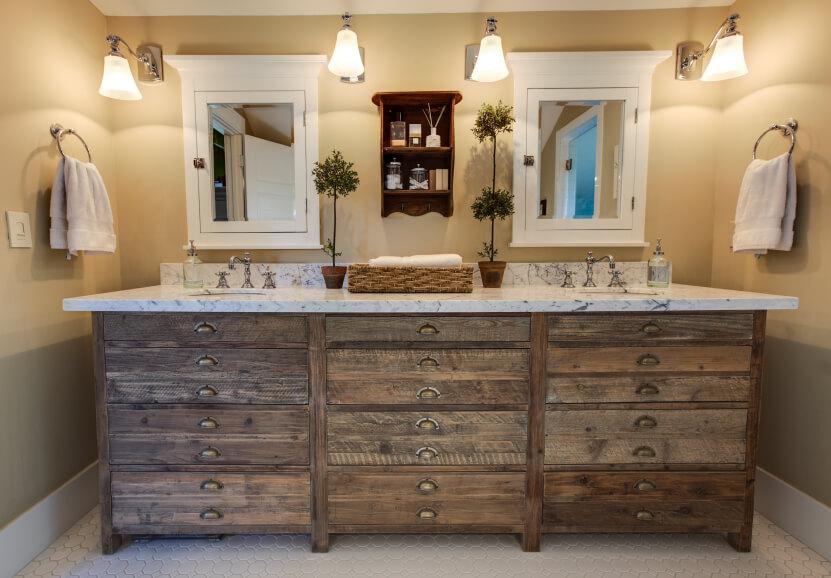} & \includegraphics[width=3cm]{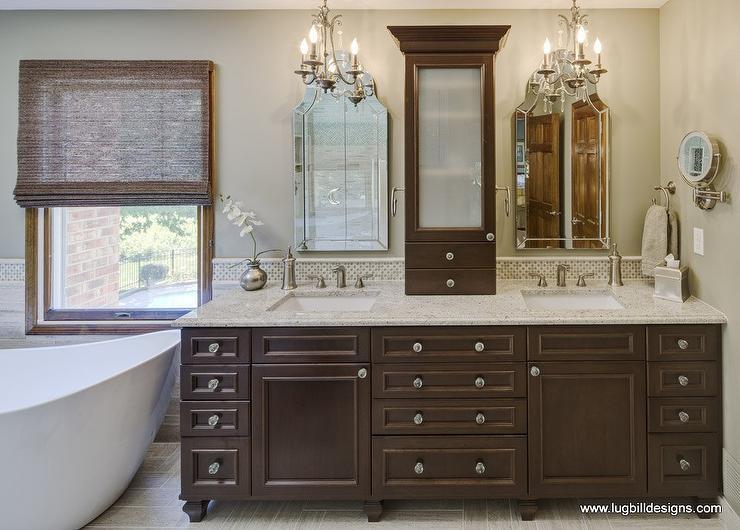} & In one of the images, the counter space between the two wash basins is empty. &  Failure Case for ViLT (Finetuned) Model, Model Prediction: False\\
\hline

\end{tabular}
\caption{Failure Cases for simple multi-modal models}
\label{tab:smul_qual_res}
\end{table*}

\begin{table*}[ht]
\centering
\begin{tabular}{|c|c|p{5cm}|p{5cm}|}
\hline
\textbf{Left Image} & \textbf{Right Image} & \textbf{Text} & \textbf{Comments} \\
\hline
\includegraphics[width=3cm]{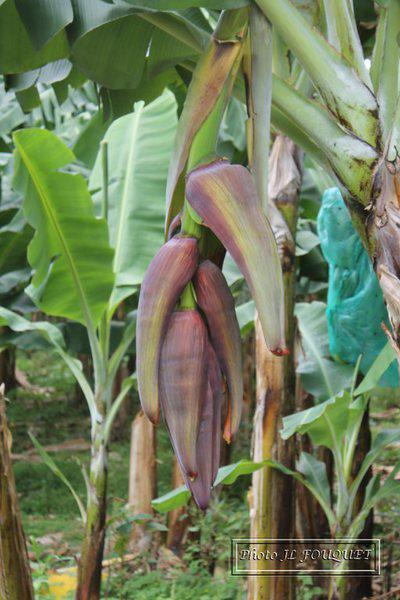} & \includegraphics[width=3cm]{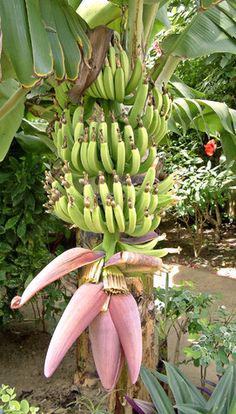} & The left and right image contains the same number of banana bunches with a purple arrow bottom. & Failure Case for VLMO Model, Model Prediction: True\\
\hline
\includegraphics[width=3cm]{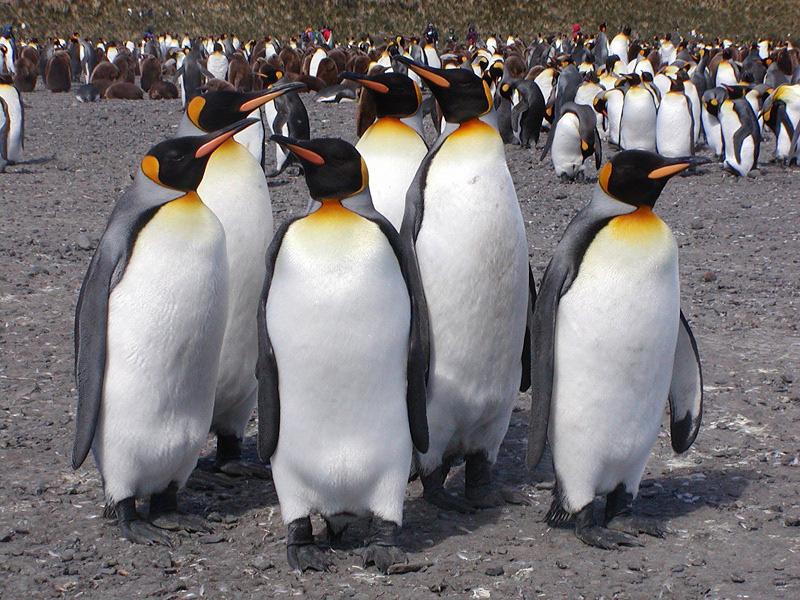} & \includegraphics[width=3cm]{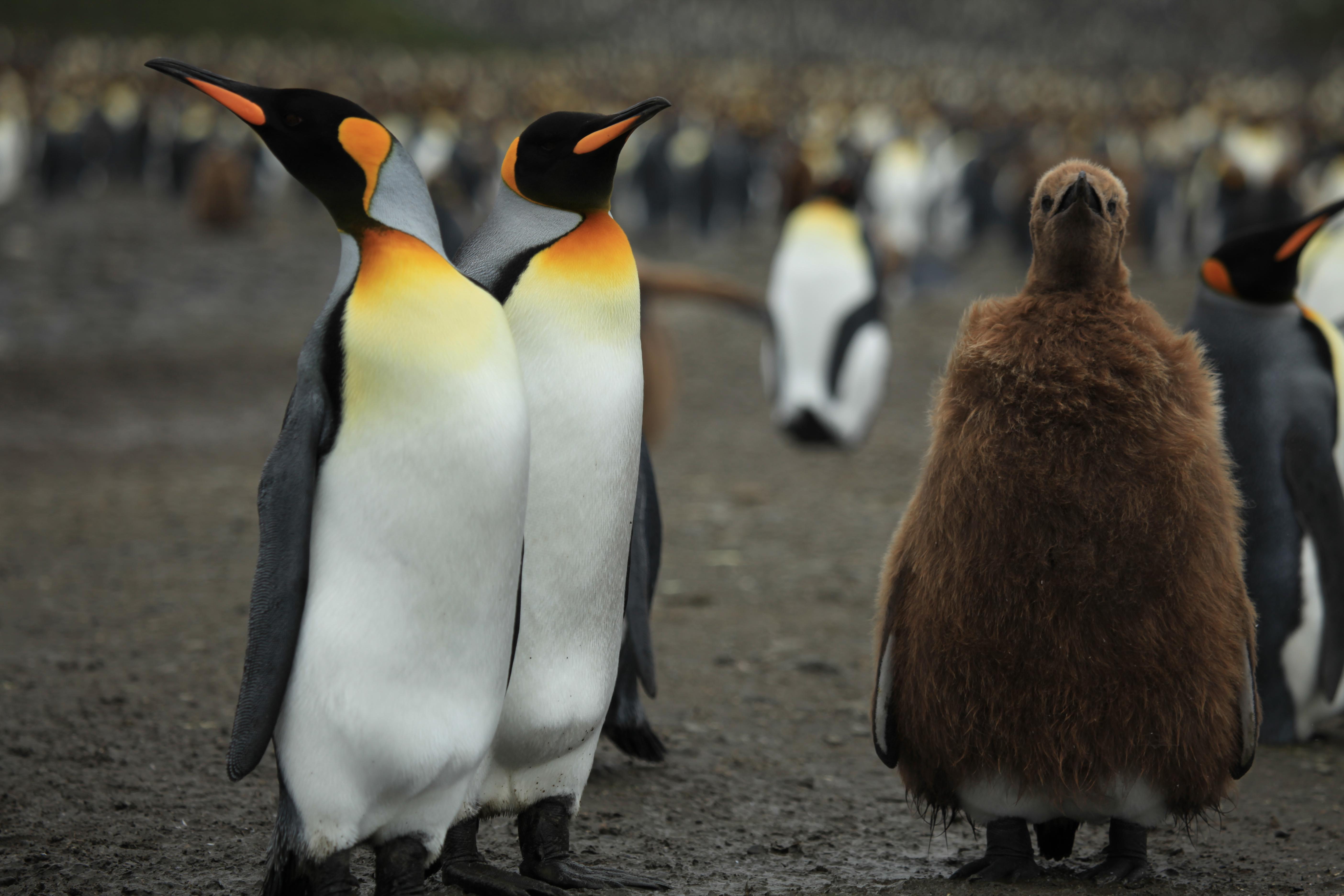} & Brown fuzzy penguins are in the background but not the foreground of the right image. & Failure Case for VLMO Model, Model Prediction: True\\
\hline
\includegraphics[width=3cm]{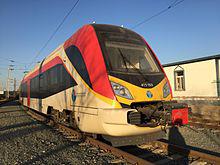} & \includegraphics[width=3cm]{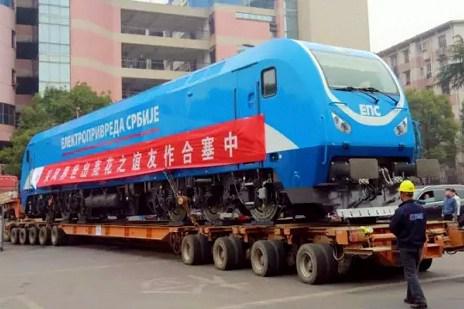} & All trains are angled rightward, and exactly one train is primarily blue while the other train has a steep sloped front and red-orange, yellow, and black accent colors. & Failure Case for X-VLM Model, Model Prediction: False\ \\
\hline
\includegraphics[width=3cm]{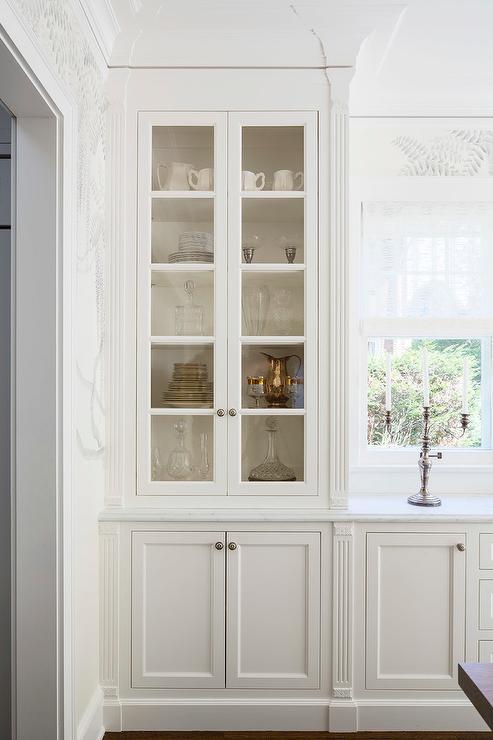} & \includegraphics[width=3cm]{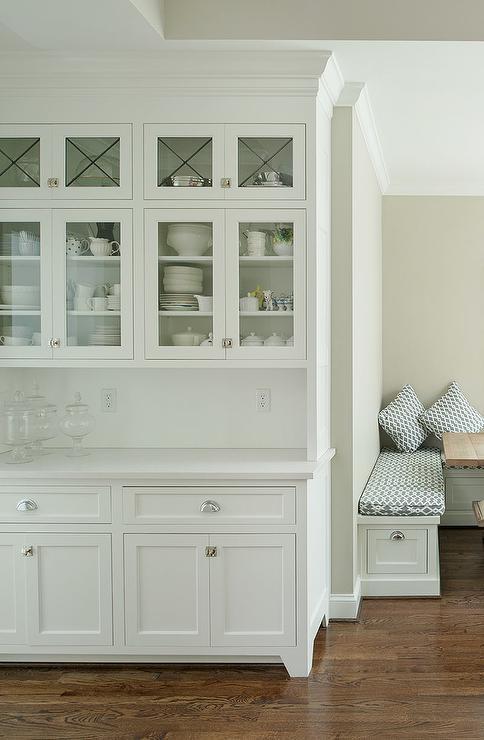} & A shelving unit covers one wall with a unique center area, but identical sections on each side, glass upper doors in one image, and solid panels in the other. & Failure Case for X-VLM Model, Model Prediction: True \\
\hline
\end{tabular}
\caption{Failure Cases for competitive multi-modal models}
\label{tab:cmul_qual_res}
\end{table*}

\begin{table*}[t]
    \centering 
    \begin{tabular}{|c|c|p{5cm}|p{5cm}|}
    \hline
\textbf{Left Image} & \textbf{Right Image} & \textbf{Text} & \textbf{Comments} \\
\hline
\includegraphics[width=3cm]{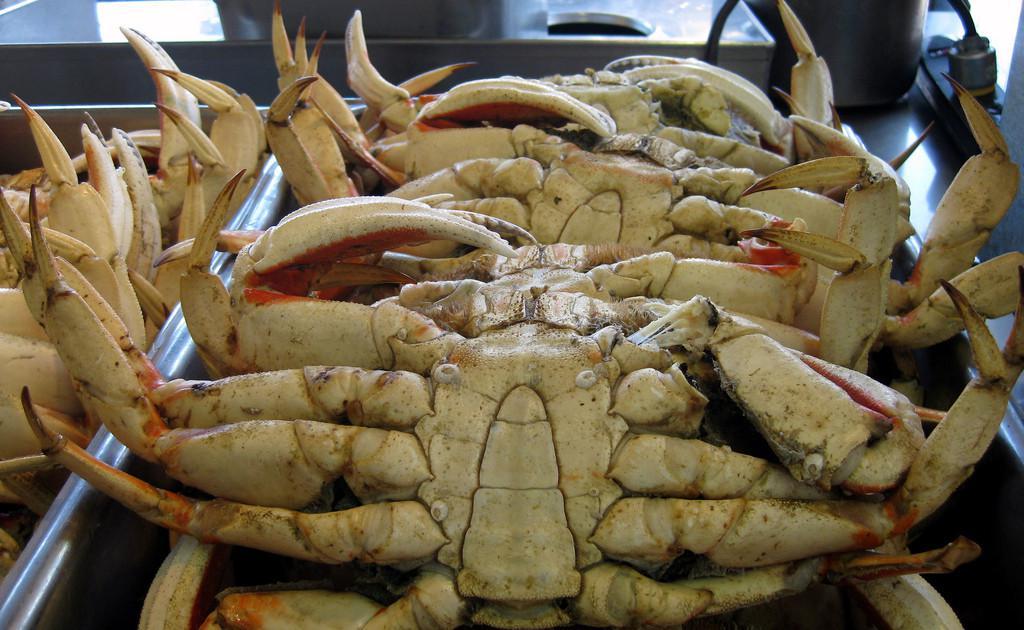}& \includegraphics[width=3cm]{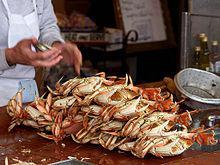} & \vspace{-2cm} Each image contains one crab facing forward, and at least one of the crabs depicted is purple-tinted. & \vspace{-2cm} Failure Case for Fromage, \qquad Model Prediction: True \\
\hline 
\includegraphics[width=3cm]{images/dev-741-1-img0.jpg}& \includegraphics[width=3cm]{images/dev-741-1-img1.jpg} & \vspace{-2cm} There are exactly two ducks in the right image. &\vspace{-2cm} Failure case for Open Flamingo, Model Prediction: False\\
\hline
\includegraphics[width=3cm]{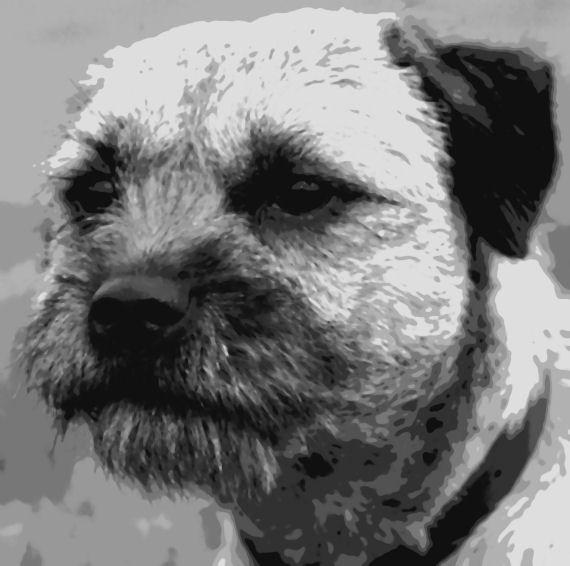} & \includegraphics[width=3cm]{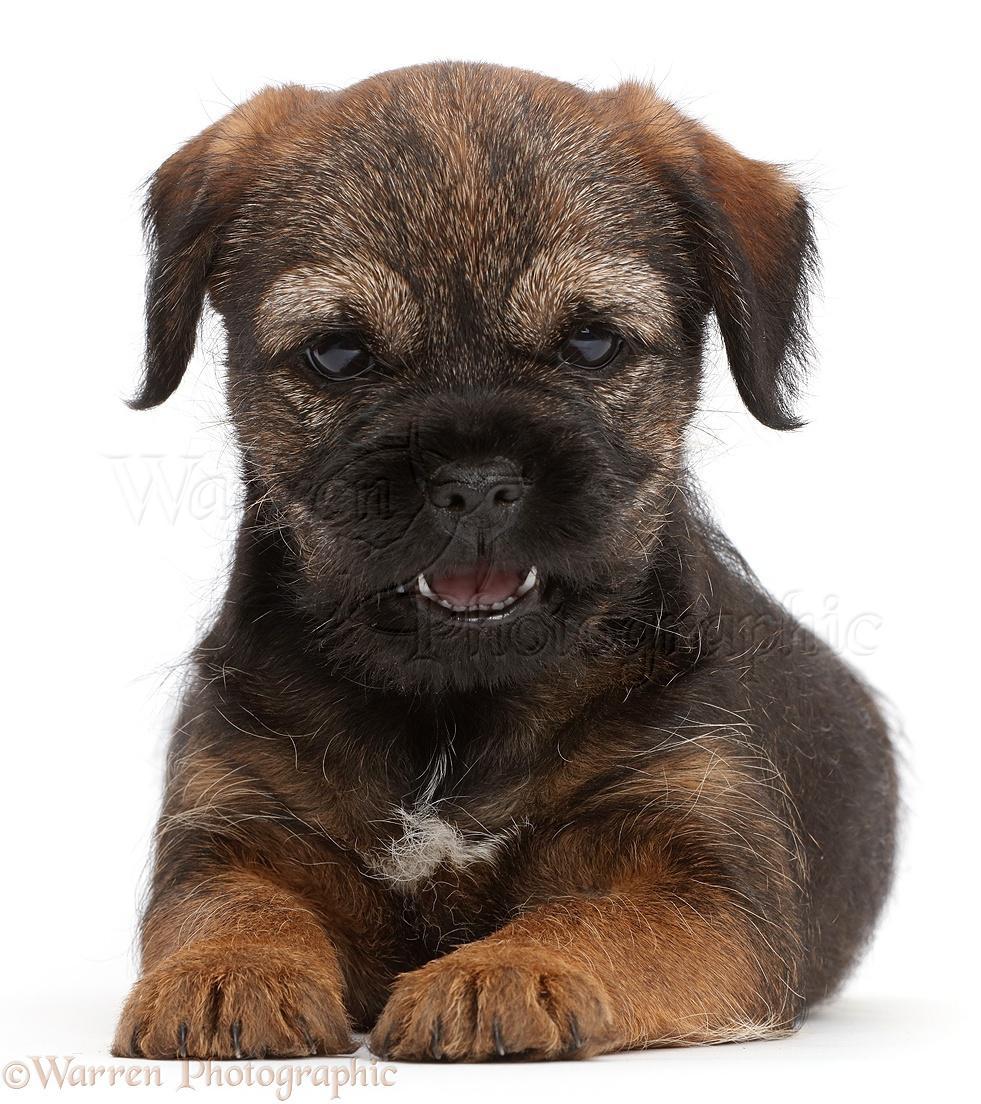} & \vspace{-3cm} The left image shows a small brown dog with its mouth open. & \vspace{-3cm} Failure case LLaVA, Model prediction: True \\
\hline
\includegraphics[width=3cm]{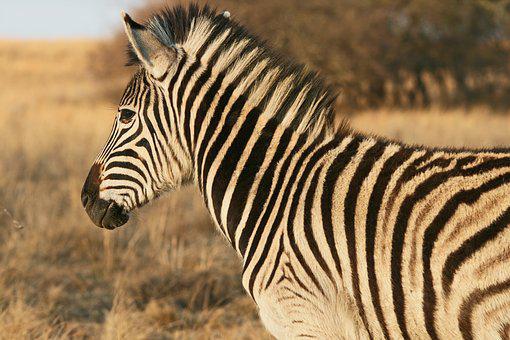} & \includegraphics[width=3cm]{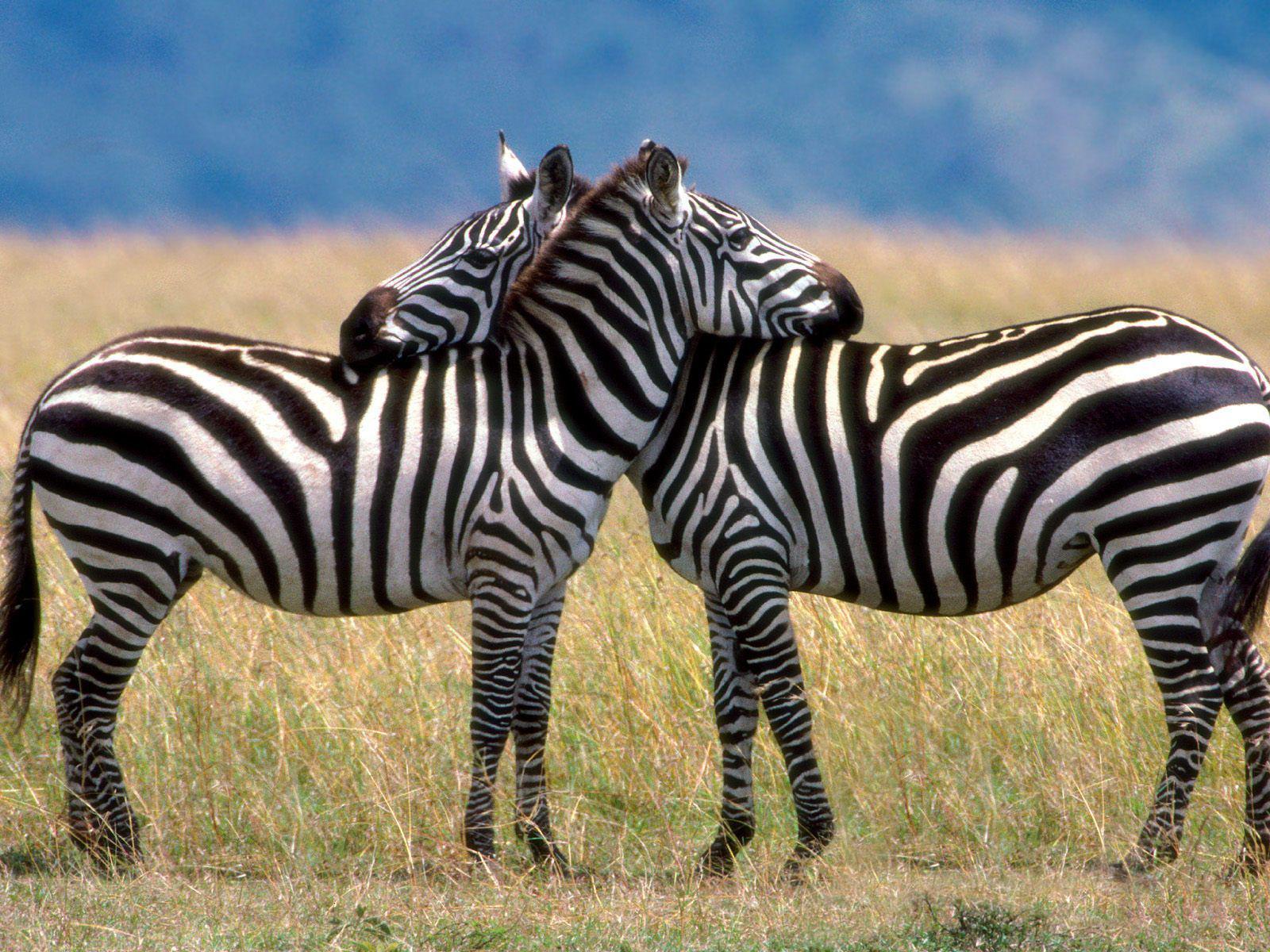} & \vspace{-2cm} One image shows a young zebra next to a leftward-facing grazing adult zebra. & \vspace{-2cm}Failure case SAM + FLAN + LimBER. Model prediction: True \\

\hline
\end{tabular}
\caption{Failure cases of Fromage and Open Flamingo}
\label{tab:bridge_fail}
\end{table*}

\begin{table*}[ht]
\centering
\begin{tabular}{|c|c|p{5cm}|p{5cm}|}
\hline
\textbf{Left Image} & \textbf{Right Image} & \textbf{Text} & \textbf{Comments} \\
\hline
\includegraphics[width=3cm]{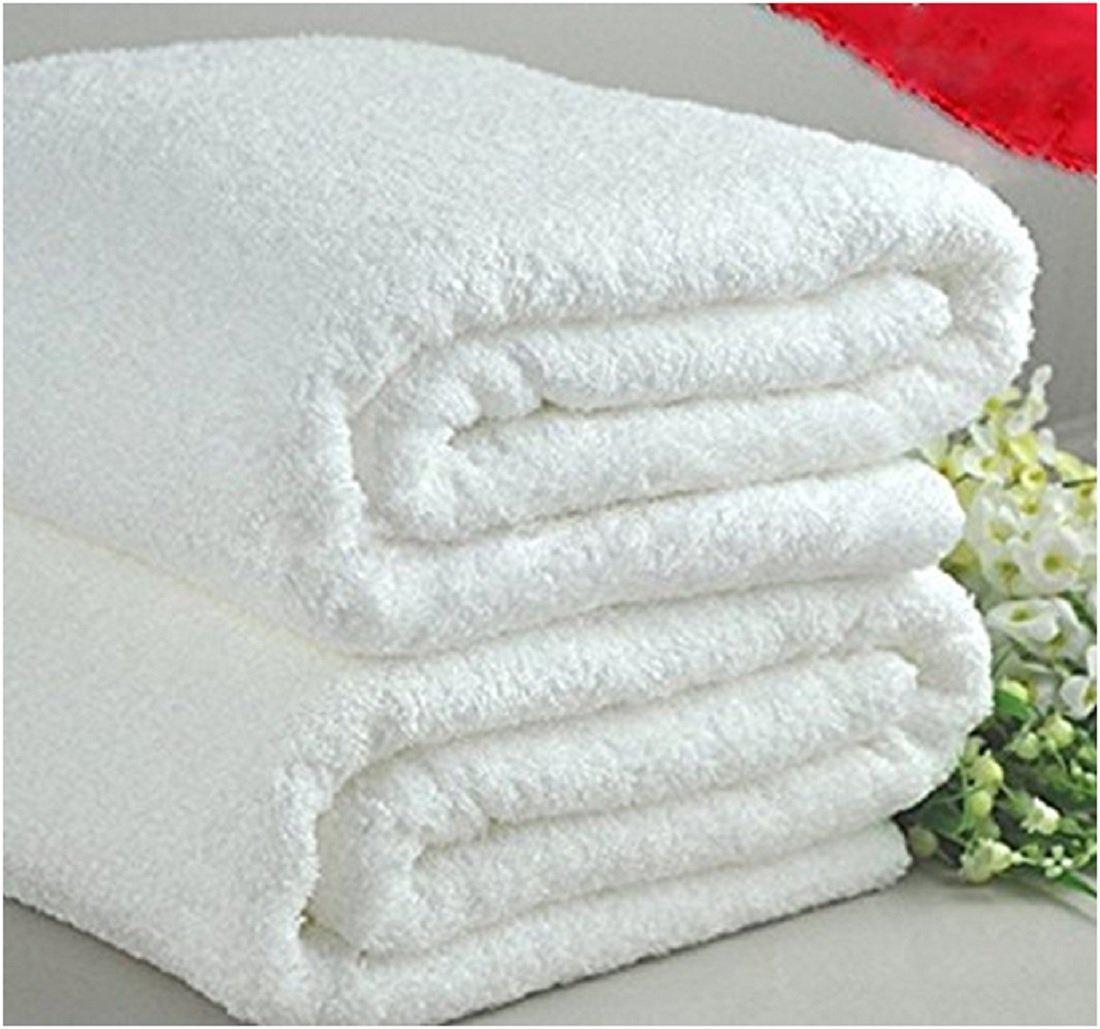} & \includegraphics[width=3cm]{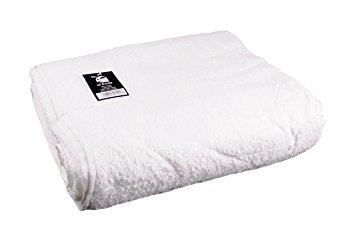} & In at least one image there is a stack of two towels with one unfolded towel. & Comparison of the three competitive multi-modal baselines, all three ViLT (Finetuned), X-VLM, VLMO get this example wrong, all three predict True whearas the actual label is False\\
\hline
\includegraphics[width=3cm]{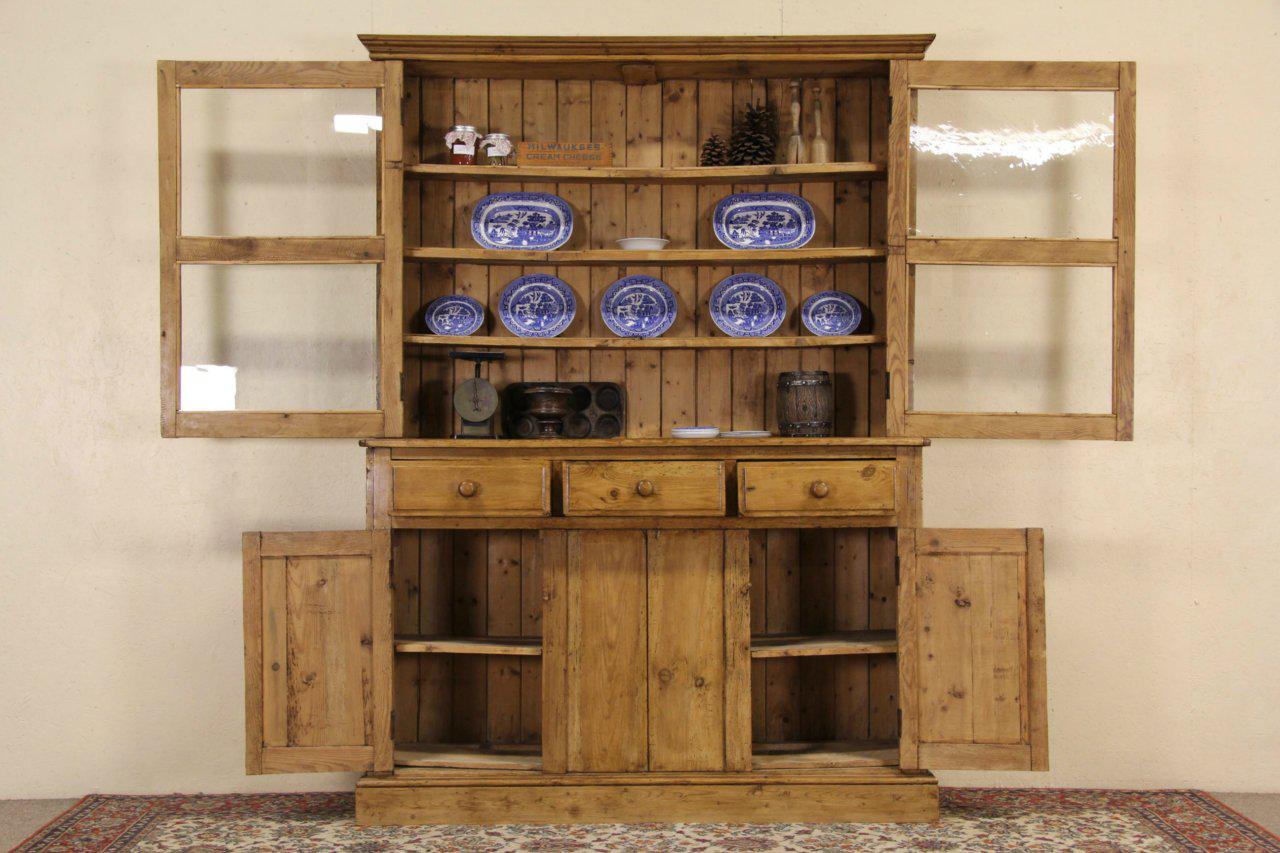} & \includegraphics[width=3cm]{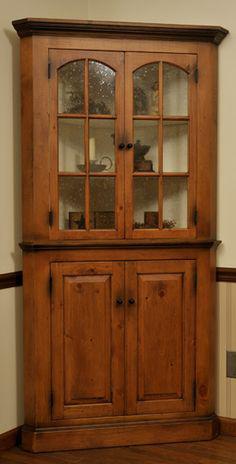} & One of the hutches does not contain any glass. & Comparison of the three competitive multi-modal baselines, ViLT (Finetuned) gets this wrong (predicts True), but X-VLM, VLMO get this example correct (predict false), actual label is False\\
\hline
\includegraphics[width=3cm]{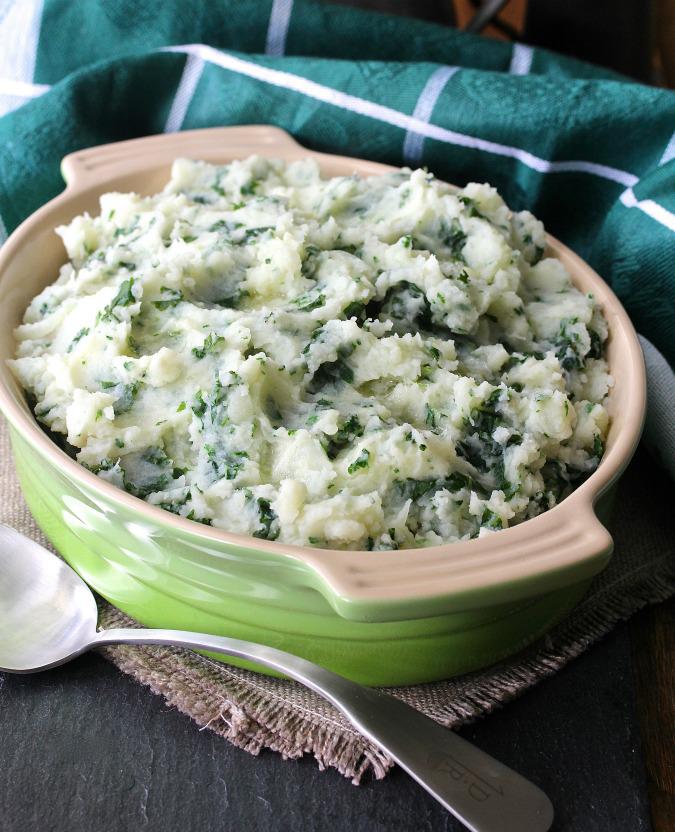} & \includegraphics[width=3cm]{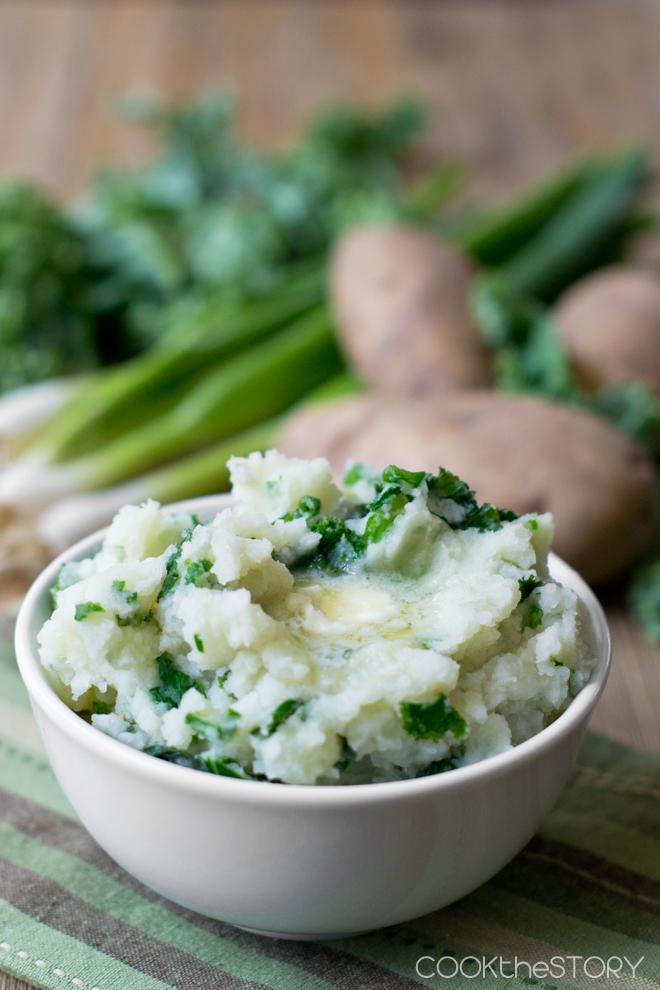} & One of the bowls is white, and one of the substances does not have any green seasoning. &  Comparison of the three competitive multi-modal baselines, ViLT (Finetuned) and VLMO get this wrong (predict True), but X-VLM gets this right, actual label is false) \\
\hline
\end{tabular}
\caption{Comparision of Failure Cases of the three competitive multi-modal baselines}
\label{tab:cmul_comp}
\end{table*}


\section{Future work and Limitations }
We provide an extensive study on various multimodal models to gauge the factors that is desired for efficient and high performance on a visio-linguistic task involving complex visual reasoning. NLVR2 is one such task that requires finegrained visual reasoning. Our evaluation on NLVR2 demonstrates that the recent trend of bridge architectures severely lack in performance when it comes to such tasks. The biggest limitation of these models is the lack of multimodal pretraining especially the image encoder. The common factor in all the models that perform very well on this task is the presence of extensive multimodal pretraining and finetuning on the train set. Due to the unavailability of A100 GPUs, we were unable to finetune the linear projection of LLaVA to compare with the corresponding linear probing numbers of ViLT and X-VLM. We plan to evaluate the most promising bridge architecture that is LLaVA by finetuning the linear projection. As researchers, our aim should be to create models that are able to perform complex visual reasoning in zero-shot without having to finetune on labeled data. To this end, we notice that LLaVA's instruction finetuning of the language model helps it to perform much better than other bridge architectures via chain of thought prompting and bootstrapping but we posit that it suffers from the lack of multimodal instruction pretraining of the image encoder. The image encoders of the model that perform well on this task are pretrained on multiple tasks which is severely lacking in LLaVA and is desired to perform well on vision-heavy task such as this. Thus we plan to validate our hypothesis that pretraining the image encoder on multimodal instruction data can improve performance on such tasks. Pretraining the whole image encoder would kind of defeat the purpose of bridge architecture and lead to an expensive pretraining procedure. So instead of pretraining the complete CLIP model, we plan to pretrain the CLIP model on multimodal instruction data via some form of CLIP Adapters similar to \citet{clipadapter}. Since bootstrapping helps, we also want to find out whether generating a chain of thought explanations from a bigger model (13B) and then finetuning a smaller model (7B) on these explanations can give better performance than zero-shot performance of a bigger model. We also aim to extend our analysis to more complex multimodal datasets such as Winoground~\cite{winoground} or CLEVR~\cite{johnson2017clevr}. LLaVA is designed to take in only one image as input and this is a limitation that we look forward to addressing in the immediate future. We found that many of the failure models of LLaVA can be attributed to this limitation. We also plan to find out if concatenating both images in NLVR2 and using multimodal models results in performance degradation. From our brief qualitative analysis of LLaVA predictions it seems like the properties of both images fuse together and the model fails on some tasks that require differentiation of these properties. This is probably because the positional encodings on the patches generated by CLIP would not be indicative of separation in the images.

\section{Ethical Concerns and Considerations}
We have used large language models such as FLAN-T5 \cite{raffel2020exploring}, OPT \cite{opt}, and GPT \cite{gpt}. While the performance of these models has been very impressive, there are some ethical considerations which need to be accounted for. First, these models can be biased \cite{bias} in certain scenarios, since the data they were trained on might be biased. In this case, the outputs of the LLM might be harmful or might exhibit implicit bias against certain sections of the society. Moreover, the training of LLMs requires an extensive amount of resources and has a very large carbon footprint. The associated environmental concerns with training such models also needs to be accounted for while using them in practice. Next, the data on which these models have been trained on might include certain aspects which were personal (such as banking data or patient data), and should not be included in training a model which is being used by the masses. Such issues can impact a lot of sections of society, especially those against whom most training datasets are biased. Other than the LLM, even the image encoders such as CLIP and SAM have been pre-trained on very large scale image data and suffer from some of the same issues such as bias, privacy concerns, and openness of the data. We believe these concerns should be a part of future work relating to this area of research. 

\clearpage
\section{Team member contributions}
\paragraph{Mrigank Raman} Running SAM+FLAN+LIMBER, helping to run LLaVA, writing analysis, future work, and proposed model

\paragraph{Kousik} Running LLaVA, getting and explaining failure modes for the bridge architectures. 

\paragraph{Pranit Chawla} Running Fromage, writing Abstract, writing introduction, runinning xvlm probing numbers

\paragraph{Mohammed Asad Karim: } Running Open-Flamingo, report writing, running and collecting ablation studies.

\bibliographystyle{acl_natbib}
\bibliography{references}

\begin{thebibliography}{58}
\expandafter\ifx\csname natexlab\endcsname\relax\def\natexlab#1{#1}\fi

\bibitem[{Alayrac et~al.(2022)Alayrac, Donahue, Luc, Miech, Barr, Hasson, Lenc,
  Mensch, Millican, Reynolds, Ring, Rutherford, Cabi, Han, Gong, Samangooei,
  Monteiro, Menick, Borgeaud, Brock, Nematzadeh, Sharifzadeh, Binkowski,
  Barreira, Vinyals, Zisserman, and Simonyan}]{open-flamingo}
Jean-Baptiste Alayrac, Jeff Donahue, Pauline Luc, Antoine Miech, Iain Barr,
  Yana Hasson, Karel Lenc, Arthur Mensch, Katie Millican, Malcolm Reynolds,
  Roman Ring, Eliza Rutherford, Serkan Cabi, Tengda Han, Zhitao Gong, Sina
  Samangooei, Marianne Monteiro, Jacob Menick, Sebastian Borgeaud, Andy Brock,
  Aida Nematzadeh, Sahand Sharifzadeh, Mikolaj Binkowski, Ricardo Barreira,
  Oriol Vinyals, Andrew Zisserman, and Karen Simonyan. 2022.
\newblock Flamingo: a visual language model for few-shot learning.
\newblock \emph{ArXiv}, abs/2204.14198.

\bibitem[{{Antol} et~al.(2015){Antol}, {Agrawal}, {Lu}, {Mitchell}, {Batra},
  {Zitnick}, and {Parikh}}]{VQA}
S.~{Antol}, A.~{Agrawal}, J.~{Lu}, M.~{Mitchell}, D.~{Batra}, C.~L. {Zitnick},
  and D.~{Parikh}. 2015.
\newblock Vqa: Visual question answering.
\newblock In \emph{2015 IEEE International Conference on Computer Vision
  (ICCV)}, pages 2425--2433.

\bibitem[{Brown et~al.(2020)Brown, Mann, Ryder, Subbiah, Kaplan, Dhariwal,
  Neelakantan, Shyam, Sastry, Askell, Agarwal, Herbert-Voss, Krueger, Henighan,
  Child, Ramesh, Ziegler, Wu, Winter, Hesse, Chen, Sigler, Litwin, Gray, Chess,
  Clark, Berner, McCandlish, Radford, Sutskever, and Amodei}]{gpt}
Tom Brown, Benjamin Mann, Nick Ryder, Melanie Subbiah, Jared~D Kaplan, Prafulla
  Dhariwal, Arvind Neelakantan, Pranav Shyam, Girish Sastry, Amanda Askell,
  Sandhini Agarwal, Ariel Herbert-Voss, Gretchen Krueger, Tom Henighan, Rewon
  Child, Aditya Ramesh, Daniel Ziegler, Jeffrey Wu, Clemens Winter, Chris
  Hesse, Mark Chen, Eric Sigler, Mateusz Litwin, Scott Gray, Benjamin Chess,
  Jack Clark, Christopher Berner, Sam McCandlish, Alec Radford, Ilya Sutskever,
  and Dario Amodei. 2020.
\newblock \href
  {https://proceedings.neurips.cc/paper/2020/file/1457c0d6bfcb4967418bfb8ac142f64a-Paper.pdf}
  {Language models are few-shot learners}.
\newblock In \emph{Advances in Neural Information Processing Systems},
  volume~33, pages 1877--1901. Curran Associates, Inc.

\bibitem[{Changpinyo et~al.(2021)Changpinyo, Sharma, Ding, and Soricut}]{cc12m}
Soravit Changpinyo, Piyush Sharma, Nan Ding, and Radu Soricut. 2021.
\newblock {Conceptual 12M}: Pushing web-scale image-text pre-training to
  recognize long-tail visual concepts.
\newblock In \emph{CVPR}.

\bibitem[{Chen et~al.(2020{\natexlab{a}})Chen, Li, Yu, El~Kholy, Ahmed, Gan,
  Cheng, and Liu}]{10.1007/978-3-030-58577-8_7}
Yen-Chun Chen, Linjie Li, Licheng Yu, Ahmed El~Kholy, Faisal Ahmed, Zhe Gan,
  Yu~Cheng, and Jingjing Liu. 2020{\natexlab{a}}.
\newblock Uniter: Universal image-text representation learning.
\newblock In \emph{Computer Vision -- ECCV 2020}, pages 104--120, Cham.
  Springer International Publishing.

\bibitem[{Chen et~al.(2020{\natexlab{b}})Chen, Li, Yu, Kholy, Ahmed, Gan,
  Cheng, and Liu}]{uniter}
Yen-Chun Chen, Linjie Li, Licheng Yu, Ahmed~El Kholy, Faisal Ahmed, Zhe Gan,
  Yu~Cheng, and Jingjing Liu. 2020{\natexlab{b}}.
\newblock Uniter: Universal image-text representation learning.
\newblock In \emph{ECCV}.

\bibitem[{Chung et~al.(2022)Chung, Hou, Longpre, Zoph, Tay, Fedus, Li, Wang,
  Dehghani, Brahma et~al.}]{chung2022scaling}
Hyung~Won Chung, Le~Hou, Shayne Longpre, Barret Zoph, Yi~Tay, William Fedus,
  Eric Li, Xuezhi Wang, Mostafa Dehghani, Siddhartha Brahma, et~al. 2022.
\newblock Scaling instruction-finetuned language models.
\newblock \emph{arXiv preprint arXiv:2210.11416}.

\bibitem[{Deng et~al.(2021)Deng, Yang, Chen, Zhou, and Li}]{transvg}
Jiajun Deng, Zhengyuan Yang, Tianlang Chen, Wengang Zhou, and Houqiang Li.
  2021.
\newblock \href {https://doi.org/10.48550/ARXIV.2104.08541} {Transvg:
  End-to-end visual grounding with transformers}.

\bibitem[{Devlin et~al.(2019{\natexlab{a}})Devlin, Chang, Lee, and
  Toutanova}]{bert}
Jacob Devlin, Ming-Wei Chang, Kenton Lee, and Kristina Toutanova.
  2019{\natexlab{a}}.
\newblock \href {https://doi.org/10.18653/v1/N19-1423} {{BERT}: Pre-training of
  deep bidirectional transformers for language understanding}.
\newblock In \emph{Proceedings of the 2019 Conference of the North {A}merican
  Chapter of the Association for Computational Linguistics: Human Language
  Technologies, Volume 1 (Long and Short Papers)}, pages 4171--4186,
  Minneapolis, Minnesota. Association for Computational Linguistics.

\bibitem[{Devlin et~al.(2019{\natexlab{b}})Devlin, Chang, Lee, and
  Toutanova}]{devlin-etal-2019-bert}
Jacob Devlin, Ming-Wei Chang, Kenton Lee, and Kristina Toutanova.
  2019{\natexlab{b}}.
\newblock \href {https://doi.org/10.18653/v1/N19-1423} {{BERT}: Pre-training of
  deep bidirectional transformers for language understanding}.
\newblock In \emph{Proceedings of the 2019 Conference of the North {A}merican
  Chapter of the Association for Computational Linguistics: Human Language
  Technologies, Volume 1 (Long and Short Papers)}, pages 4171--4186,
  Minneapolis, Minnesota. Association for Computational Linguistics.

\bibitem[{Dosovitskiy et~al.(2021)Dosovitskiy, Beyer, Kolesnikov, Weissenborn,
  Zhai, Unterthiner, Dehghani, Minderer, Heigold, Gelly, Uszkoreit, and
  Houlsby}]{vit}
Alexey Dosovitskiy, Lucas Beyer, Alexander Kolesnikov, Dirk Weissenborn,
  Xiaohua Zhai, Thomas Unterthiner, Mostafa Dehghani, Matthias Minderer, Georg
  Heigold, Sylvain Gelly, Jakob Uszkoreit, and Neil Houlsby. 2021.
\newblock \href {https://openreview.net/forum?id=YicbFdNTTy} {An image is worth
  16x16 words: Transformers for image recognition at scale}.
\newblock In \emph{International Conference on Learning Representations}.

\bibitem[{Eichenberg et~al.(2021)Eichenberg, Black, Weinbach, Parcalabescu, and
  Frank}]{magma}
Constantin Eichenberg, Sidney Black, Samuel Weinbach, Letitia Parcalabescu, and
  Anette Frank. 2021.
\newblock \href {https://doi.org/10.48550/ARXIV.2112.05253} {Magma --
  multimodal augmentation of generative models through adapter-based
  finetuning}.

\bibitem[{Gao et~al.(2021)Gao, Geng, Zhang, Ma, Fang, Zhang, Li, and
  Qiao}]{clipadapter}
Peng Gao, Shijie Geng, Renrui Zhang, Teli Ma, Rongyao Fang, Yongfeng Zhang,
  Hongsheng Li, and Yu~Qiao. 2021.
\newblock Clip-adapter: Better vision-language models with feature adapters.
\newblock \emph{arXiv preprint arXiv:2110.04544}.

\bibitem[{{Girshick}(2015)}]{fast_rcnn}
R.~{Girshick}. 2015.
\newblock Fast r-cnn.
\newblock In \emph{2015 IEEE International Conference on Computer Vision
  (ICCV)}, pages 1440--1448.

\bibitem[{Goyal et~al.(2022)Goyal, Kumar, Garg, Kolter, and
  Raghunathan}]{sachin}
Sachin Goyal, Ananya Kumar, Sankalp Garg, Zico Kolter, and Aditi Raghunathan.
  2022.
\newblock Finetune like you pretrain: Improved finetuning of zero-shot vision
  models.
\newblock \emph{arXiv preprint arXiv:2212.00638}.

\bibitem[{Goyal et~al.(2016)Goyal, Khot, Summers{-}Stay, Batra, and
  Parikh}]{GoyalKSBP16}
Yash Goyal, Tejas Khot, Douglas Summers{-}Stay, Dhruv Batra, and Devi Parikh.
  2016.
\newblock \href {http://arxiv.org/abs/1612.00837} {Making the {V} in {VQA}
  matter: Elevating the role of image understanding in visual question
  answering}.
\newblock \emph{CoRR}, abs/1612.00837.

\bibitem[{Halawani et~al.(2006)Halawani, Teynor, Setia, Brunner, and
  Burkhardt}]{applications_image_retrieval}
Alaa Halawani, Alexandra Teynor, Lokesh Setia, Gerd Brunner, and Hans
  Burkhardt. 2006.
\newblock Fundamentals and applications of image retrieval: An overview.
\newblock \emph{Datenbank-Spektrum}, 18:14--23.

\bibitem[{He et~al.(2022)He, Chen, Xie, Li, Doll{\'a}r, and Girshick}]{mae}
Kaiming He, Xinlei Chen, Saining Xie, Yanghao Li, Piotr Doll{\'a}r, and Ross
  Girshick. 2022.
\newblock Masked autoencoders are scalable vision learners.
\newblock In \emph{Proceedings of the IEEE/CVF Conference on Computer Vision
  and Pattern Recognition}, pages 16000--16009.

\bibitem[{He et~al.(2016)He, Zhang, Ren, and Sun}]{resnet}
Kaiming He, Xiangyu Zhang, Shaoqing Ren, and Jian Sun. 2016.
\newblock Deep residual learning for image recognition.
\newblock In \emph{Proceedings of the IEEE Conference on Computer Vision and
  Pattern Recognition (CVPR)}.

\bibitem[{Huang et~al.(2022)Huang, Gu, Hou, Wu, Wang, Yu, and
  Han}]{selfimprove}
Jiaxin Huang, Shixiang~Shane Gu, Le~Hou, Yuexin Wu, Xuezhi Wang, Hongkun Yu,
  and Jiawei Han. 2022.
\newblock Large language models can self-improve.
\newblock \emph{arXiv preprint arXiv:2210.11610}.

\bibitem[{Huang et~al.(2021)Huang, Zeng, Huang, Liu, Fu, and
  Fu}]{Huang_2021_CVPR}
Zhicheng Huang, Zhaoyang Zeng, Yupan Huang, Bei Liu, Dongmei Fu, and Jianlong
  Fu. 2021.
\newblock Seeing out of the box: End-to-end pre-training for vision-language
  representation learning.
\newblock In \emph{Proceedings of the IEEE/CVF Conference on Computer Vision
  and Pattern Recognition (CVPR)}, pages 12976--12985.

\bibitem[{Johnson et~al.(2017)Johnson, Hariharan, Van Der~Maaten, Fei-Fei,
  Lawrence~Zitnick, and Girshick}]{johnson2017clevr}
Justin Johnson, Bharath Hariharan, Laurens Van Der~Maaten, Li~Fei-Fei,
  C~Lawrence~Zitnick, and Ross Girshick. 2017.
\newblock Clevr: A diagnostic dataset for compositional language and elementary
  visual reasoning.
\newblock In \emph{Proceedings of the IEEE conference on computer vision and
  pattern recognition}, pages 2901--2910.

\bibitem[{Kamath et~al.(2021)Kamath, Singh, LeCun, Synnaeve, Misra, and
  Carion}]{mdetr}
Aishwarya Kamath, Mannat Singh, Yann LeCun, Gabriel Synnaeve, Ishan Misra, and
  Nicolas Carion. 2021.
\newblock \href {https://doi.org/10.48550/ARXIV.2104.12763} {Mdetr -- modulated
  detection for end-to-end multi-modal understanding}.

\bibitem[{Karpathy and Fei-Fei(2015)}]{caption1}
Andrej Karpathy and Li~Fei-Fei. 2015.
\newblock Deep visual-semantic alignments for generating image descriptions.
\newblock In \emph{Proceedings of the IEEE conference on computer vision and
  pattern recognition}, pages 3128--3137.

\bibitem[{Kim et~al.(2021)Kim, Son, and Kim}]{vilt}
Wonjae Kim, Bokyung Son, and Ildoo Kim. 2021.
\newblock \href {https://doi.org/10.48550/ARXIV.2102.03334} {Vilt:
  Vision-and-language transformer without convolution or region supervision}.

\bibitem[{Kirillov et~al.(2023)Kirillov, Mintun, Ravi, Mao, Rolland, Gustafson,
  Xiao, Whitehead, Berg, Lo et~al.}]{sam}
Alexander Kirillov, Eric Mintun, Nikhila Ravi, Hanzi Mao, Chloe Rolland, Laura
  Gustafson, Tete Xiao, Spencer Whitehead, Alexander~C Berg, Wan-Yen Lo, et~al.
  2023.
\newblock Segment anything.
\newblock \emph{arXiv preprint arXiv:2304.02643}.

\bibitem[{Koh et~al.(2023)Koh, Salakhutdinov, and Fried}]{fromage}
Jing~Yu Koh, Ruslan Salakhutdinov, and Daniel Fried. 2023.
\newblock \href {https://doi.org/10.48550/ARXIV.2301.13823} {Grounding language
  models to images for multimodal generation}.

\bibitem[{Li et~al.(2021)Li, Selvaraju, Gotmare, Joty, Xiong, and
  Hoi}]{NEURIPS2021_50525975}
Junnan Li, Ramprasaath Selvaraju, Akhilesh Gotmare, Shafiq Joty, Caiming Xiong,
  and Steven Chu~Hong Hoi. 2021.
\newblock \href
  {https://proceedings.neurips.cc/paper/2021/file/505259756244493872b7709a8a01b536-Paper.pdf}
  {Align before fuse: Vision and language representation learning with momentum
  distillation}.
\newblock In \emph{Advances in Neural Information Processing Systems},
  volume~34, pages 9694--9705. Curran Associates, Inc.

\bibitem[{Li et~al.(2019)Li, Yatskar, Yin, Hsieh, and Chang}]{visbert}
Liunian~Harold Li, Mark Yatskar, Da~Yin, Cho{-}Jui Hsieh, and Kai{-}Wei Chang.
  2019.
\newblock \href {http://arxiv.org/abs/1908.03557} {Visualbert: {A} simple and
  performant baseline for vision and language}.
\newblock \emph{CoRR}, abs/1908.03557.

\bibitem[{Lin et~al.(2015)Lin, Maire, Belongie, Bourdev, Girshick, Hays,
  Perona, Ramanan, Zitnick, and Dollár}]{mscoco}
Tsung-Yi Lin, Michael Maire, Serge Belongie, Lubomir Bourdev, Ross Girshick,
  James Hays, Pietro Perona, Deva Ramanan, C.~Lawrence Zitnick, and Piotr
  Dollár. 2015.
\newblock \href {http://arxiv.org/abs/1405.0312} {Microsoft coco: Common
  objects in context}.

\bibitem[{Liu et~al.(2021)Liu, Bugliarello, Ponti, Reddy, Collier, and
  Elliott}]{marvl}
Fangyu Liu, Emanuele Bugliarello, Edoardo~Maria Ponti, Siva Reddy, Nigel
  Collier, and Desmond Elliott. 2021.
\newblock \href {http://arxiv.org/abs/2109.13238} {Visually grounded reasoning
  across languages and cultures}.
\newblock \emph{CoRR}, abs/2109.13238.

\bibitem[{Liu et~al.(2023)Liu, Li, Wu, and Lee}]{llava}
Haotian Liu, Chunyuan Li, Qingyang Wu, and Yong~Jae Lee. 2023.
\newblock Visual instruction tuning.
\newblock \emph{arXiv preprint arXiv:2304.08485}.

\bibitem[{Liu et~al.(2019)Liu, Ott, Goyal, Du, Joshi, Chen, Levy, Lewis,
  Zettlemoyer, and Stoyanov}]{roberta}
Yinhan Liu, Myle Ott, Naman Goyal, Jingfei Du, Mandar Joshi, Danqi Chen, Omer
  Levy, Mike Lewis, Luke Zettlemoyer, and Veselin Stoyanov. 2019.
\newblock Roberta: A robustly optimized bert pretraining approach.
\newblock \emph{ArXiv}, abs/1907.11692.

\bibitem[{Lu et~al.(2019)Lu, Batra, Parikh, and Lee}]{vilbert}
Jiasen Lu, Dhruv Batra, Devi Parikh, and Stefan Lee. 2019.
\newblock Vilbert: Pretraining task-agnostic visiolinguistic representations
  for vision-and-language tasks.
\newblock In \emph{Advances in Neural Information Processing Systems}, pages
  13--23.

\bibitem[{Meng et~al.(2022)Meng, Sharma, Andonian, Belinkov, and Bau}]{edit}
Kevin Meng, Arnab~Sen Sharma, Alex Andonian, Yonatan Belinkov, and David Bau.
  2022.
\newblock \href {https://doi.org/10.48550/ARXIV.2210.07229} {Mass-editing
  memory in a transformer}.

\bibitem[{Merullo et~al.(2022)Merullo, Castricato, Eickhoff, and
  Pavlick}]{limber}
Jack Merullo, Louis Castricato, Carsten Eickhoff, and Ellie Pavlick. 2022.
\newblock \href {https://doi.org/10.48550/ARXIV.2209.15162} {Linearly mapping
  from image to text space}.

\bibitem[{Radford et~al.(2021)Radford, Kim, Hallacy, Ramesh, Goh, Agarwal,
  Sastry, Askell, Mishkin, Clark, Krueger, and Sutskever}]{clip}
Alec Radford, Jong~Wook Kim, Chris Hallacy, Aditya Ramesh, Gabriel Goh,
  Sandhini Agarwal, Girish Sastry, Amanda Askell, Pamela Mishkin, Jack Clark,
  Gretchen Krueger, and Ilya Sutskever. 2021.
\newblock \href {http://arxiv.org/abs/2103.00020} {Learning transferable visual
  models from natural language supervision}.
\newblock \emph{CoRR}, abs/2103.00020.

\bibitem[{Radford et~al.(2019)Radford, Wu, Child, Luan, Amodei, and
  Sutskever}]{Radford2019LanguageMA}
Alec Radford, Jeff Wu, Rewon Child, David Luan, Dario Amodei, and Ilya
  Sutskever. 2019.
\newblock Language models are unsupervised multitask learners.

\bibitem[{Raffel et~al.(2020)Raffel, Shazeer, Roberts, Lee, Narang, Matena,
  Zhou, Li, and Liu}]{raffel2020exploring}
Colin Raffel, Noam Shazeer, Adam Roberts, Katherine Lee, Sharan Narang, Michael
  Matena, Yanqi Zhou, Wei Li, and Peter~J Liu. 2020.
\newblock Exploring the limits of transfer learning with a unified text-to-text
  transformer.
\newblock \emph{The Journal of Machine Learning Research}, 21(1):5485--5551.

\bibitem[{Ren et~al.(2017)Ren, He, Girshick, and Sun}]{fasterrcnn}
Shaoqing Ren, Kaiming He, Ross Girshick, and Jian Sun. 2017.
\newblock \href {https://doi.org/10.1109/TPAMI.2016.2577031} {Faster r-cnn:
  Towards real-time object detection with region proposal networks}.
\newblock \emph{IEEE Transactions on Pattern Analysis and Machine
  Intelligence}, 39(6):1137--1149.

\bibitem[{Schuhmann et~al.(2022)Schuhmann, Beaumont, Vencu, Gordon, Wightman,
  Cherti, Coombes, Katta, Mullis, Wortsman et~al.}]{laion}
Christoph Schuhmann, Romain Beaumont, Richard Vencu, Cade Gordon, Ross
  Wightman, Mehdi Cherti, Theo Coombes, Aarush Katta, Clayton Mullis, Mitchell
  Wortsman, et~al. 2022.
\newblock Laion-5b: An open large-scale dataset for training next generation
  image-text models.
\newblock \emph{arXiv preprint arXiv:2210.08402}.

\bibitem[{Sharma et~al.(2018)Sharma, Ding, Goodman, and Soricut}]{cc3m}
Piyush Sharma, Nan Ding, Sebastian Goodman, and Radu Soricut. 2018.
\newblock \href {https://doi.org/10.18653/v1/P18-1238} {Conceptual captions: A
  cleaned, hypernymed, image alt-text dataset for automatic image captioning}.
\newblock In \emph{Proceedings of the 56th Annual Meeting of the Association
  for Computational Linguistics (Volume 1: Long Papers)}, pages 2556--2565,
  Melbourne, Australia. Association for Computational Linguistics.

\bibitem[{Suhr et~al.(2017{\natexlab{a}})Suhr, Lewis, Yeh, and
  Artzi}]{suhr2017corpus}
Alane Suhr, Mike Lewis, James Yeh, and Yoav Artzi. 2017{\natexlab{a}}.
\newblock A corpus of natural language for visual reasoning.
\newblock In \emph{Proceedings of the 55th Annual Meeting of the Association
  for Computational Linguistics (Volume 2: Short Papers)}, pages 217--223.

\bibitem[{Suhr et~al.(2017{\natexlab{b}})Suhr, Lewis, Yeh, and
  Artzi}]{suhr-etal-2017-corpus}
Alane Suhr, Mike Lewis, James Yeh, and Yoav Artzi. 2017{\natexlab{b}}.
\newblock \href {https://doi.org/10.18653/v1/P17-2034} {A corpus of natural
  language for visual reasoning}.
\newblock In \emph{Proceedings of the 55th Annual Meeting of the Association
  for Computational Linguistics (Volume 2: Short Papers)}, pages 217--223,
  Vancouver, Canada. Association for Computational Linguistics.

\bibitem[{Suhr et~al.(2018{\natexlab{a}})Suhr, Zhou, Zhang, Zhang, Bai, and
  Artzi}]{suhr2018corpus}
Alane Suhr, Stephanie Zhou, Ally Zhang, Iris Zhang, Huajun Bai, and Yoav Artzi.
  2018{\natexlab{a}}.
\newblock A corpus for reasoning about natural language grounded in
  photographs.
\newblock \emph{arXiv preprint arXiv:1811.00491}.

\bibitem[{Suhr et~al.(2018{\natexlab{b}})Suhr, Zhou, Zhang, Bai, and
  Artzi}]{nlvr2}
Alane Suhr, Stephanie Zhou, Iris Zhang, Huajun Bai, and Yoav Artzi.
  2018{\natexlab{b}}.
\newblock \href {http://arxiv.org/abs/1811.00491} {A corpus for reasoning about
  natural language grounded in photographs}.
\newblock \emph{CoRR}, abs/1811.00491.

\bibitem[{Talboy and Fuller(2023)}]{bias}
Alaina~N. Talboy and Elizabeth Fuller. 2023.
\newblock \href {http://arxiv.org/abs/2304.01358} {Challenging the appearance
  of machine intelligence: Cognitive bias in llms}.

\bibitem[{Tan and Bansal(2019)}]{lxmert}
Hao Tan and Mohit Bansal. 2019.
\newblock Lxmert: Learning cross-modality encoder representations from
  transformers.
\newblock In \emph{Proceedings of the 2019 Conference on Empirical Methods in
  Natural Language Processing}.

\bibitem[{Thrush et~al.(2022)Thrush, Jiang, Bartolo, Singh, Williams, Kiela,
  and Ross}]{winoground}
Tristan Thrush, Ryan Jiang, Max Bartolo, Amanpreet Singh, Adina Williams, Douwe
  Kiela, and Candace Ross. 2022.
\newblock Winoground: Probing vision and language models for visio-linguistic
  compositionality.
\newblock In \emph{Proceedings of the IEEE/CVF Conference on Computer Vision
  and Pattern Recognition}, pages 5238--5248.

\bibitem[{Touvron et~al.(2023)Touvron, Lavril, Izacard, Martinet, Lachaux,
  Lacroix, Rozière, Goyal, Hambro, Azhar, Rodriguez, Joulin, Grave, and
  Lample}]{llama}
Hugo Touvron, Thibaut Lavril, Gautier Izacard, Xavier Martinet, Marie-Anne
  Lachaux, Timothée Lacroix, Baptiste Rozière, Naman Goyal, Eric Hambro,
  Faisal Azhar, Aurelien Rodriguez, Armand Joulin, Edouard Grave, and Guillaume
  Lample. 2023.
\newblock \href {https://doi.org/10.48550/ARXIV.2302.13971} {Llama: Open and
  efficient foundation language models}.

\bibitem[{Tsimpoukelli et~al.(2021)Tsimpoukelli, Menick, Cabi, Eslami, Vinyals,
  and Hill}]{frozen}
Maria Tsimpoukelli, Jacob Menick, Serkan Cabi, S.~M.~Ali Eslami, Oriol Vinyals,
  and Felix Hill. 2021.
\newblock \href {https://doi.org/10.48550/ARXIV.2106.13884} {Multimodal
  few-shot learning with frozen language models}.

\bibitem[{Wang et~al.(2021{\natexlab{a}})Wang, Bao, Dong, and Wei}]{vlmo}
Wenhui Wang, Hangbo Bao, Li~Dong, and Furu Wei. 2021{\natexlab{a}}.
\newblock \href {http://arxiv.org/abs/2111.02358} {Vlmo: Unified
  vision-language pre-training with mixture-of-modality-experts}.
\newblock \emph{CoRR}, abs/2111.02358.

\bibitem[{Wang et~al.(2021{\natexlab{b}})Wang, Yu, Yu, Dai, Tsvetkov, and
  Cao}]{Wang2021SimVLMSV}
Zirui Wang, Jiahui Yu, Adams~Wei Yu, Zihang Dai, Yulia Tsvetkov, and Yuan Cao.
  2021{\natexlab{b}}.
\newblock Simvlm: Simple visual language model pretraining with weak
  supervision.
\newblock \emph{ArXiv}, abs/2108.10904.

\bibitem[{Wei et~al.(2022{\natexlab{a}})Wei, Wang, Schuurmans, Bosma, Chi, Le,
  and Zhou}]{cot}
Jason Wei, Xuezhi Wang, Dale Schuurmans, Maarten Bosma, Ed~H. Chi, Quoc Le, and
  Denny Zhou. 2022{\natexlab{a}}.
\newblock \href {http://arxiv.org/abs/2201.11903} {Chain of thought prompting
  elicits reasoning in large language models}.
\newblock \emph{CoRR}, abs/2201.11903.

\bibitem[{Wei et~al.(2022{\natexlab{b}})Wei, Wang, Schuurmans, Bosma, Ichter,
  Xia, Chi, Le, and Zhou}]{chain_thought}
Jason Wei, Xuezhi Wang, Dale Schuurmans, Maarten Bosma, Brian Ichter, Fei Xia,
  Ed~Chi, Quoc Le, and Denny Zhou. 2022{\natexlab{b}}.
\newblock \href {https://doi.org/10.48550/ARXIV.2201.11903} {Chain-of-thought
  prompting elicits reasoning in large language models}.

\bibitem[{Zeng et~al.(2021)Zeng, Zhang, and Li}]{xvlm}
Yan Zeng, Xinsong Zhang, and Hang Li. 2021.
\newblock \href {http://arxiv.org/abs/2111.08276} {Multi-grained vision
  language pre-training: Aligning texts with visual concepts}.
\newblock \emph{CoRR}, abs/2111.08276.

\bibitem[{Zhang et~al.(2022)Zhang, Roller, Goyal, Artetxe, Chen, Chen, Dewan,
  Diab, Li, Lin, Mihaylov, Ott, Shleifer, Shuster, Simig, Koura, Sridhar, Wang,
  and Zettlemoyer}]{opt}
Susan Zhang, Stephen Roller, Naman Goyal, Mikel Artetxe, Moya Chen, Shuohui
  Chen, Christopher Dewan, Mona Diab, Xian Li, Xi~Victoria Lin, Todor Mihaylov,
  Myle Ott, Sam Shleifer, Kurt Shuster, Daniel Simig, Punit~Singh Koura, Anjali
  Sridhar, Tianlu Wang, and Luke Zettlemoyer. 2022.
\newblock \href {https://doi.org/10.48550/ARXIV.2205.01068} {Opt: Open
  pre-trained transformer language models}.

\bibitem[{Zhuang et~al.(2021)Zhuang, Wayne, Ya, and
  Jun}]{zhuang-etal-2021-robustly}
Liu Zhuang, Lin Wayne, Shi Ya, and Zhao Jun. 2021.
\newblock \href {https://aclanthology.org/2021.ccl-1.108} {A robustly optimized
  {BERT} pre-training approach with post-training}.
\newblock In \emph{Proceedings of the 20th Chinese National Conference on
  Computational Linguistics}, pages 1218--1227, Huhhot, China. Chinese
  Information Processing Society of China.

\end{thebibliography}


\end{document}